\documentclass[10pt,journal,compsoc]{IEEEtran}
\ifCLASSOPTIONcompsoc
  \usepackage[nocompress]{cite}
\else
  \usepackage{cite}
\fi
\ifCLASSINFOpdf
   \usepackage[pdftex]{graphicx}
\else
\fi
\usepackage{amsthm,amsmath,amssymb}
\usepackage{mathrsfs}
\usepackage{booktabs}
\usepackage{multirow}
\usepackage{multicol}
\usepackage{xcolor}
\usepackage{color}
\usepackage{hyperref}

\newcommand{\tabincell}[2]{\begin{tabular}{@{}#1@{}}#2\end{tabular}}

\begin{document}
%
\title{Efficient Multimodal Transformer with Dual-Level Feature Restoration for Robust Multimodal Sentiment Analysis}
%
%
%
%

\author{Licai~Sun,
        Zheng~Lian,
        Bin~Liu,~\IEEEmembership{Member,~IEEE,}
        and~Jianhua~Tao,~\IEEEmembership{Senior~Member,~IEEE}
\IEEEcompsocitemizethanks{
\IEEEcompsocthanksitem Licai Sun is with the School of Artificial Intelligence, University of Chinese Academy of Sciences, Beijing, China, 100049, and the National Laboratory of Pattern Recognition, Institute of Automation, Chinese Academy of Sciences, Beijing, China, 100190.\protect\\
E-mail: sunlicai2019@ia.ac.cn
\IEEEcompsocthanksitem Zheng Lian and Bin Liu are with the National Laboratory of Pattern Recognition, Institute of Automation, Chinese Academy of Sciences, Beijing, China, 100190.\protect\\
E-mail: lianzheng2016@ia.ac.cn, liubin@nlpr.ia.ac.cn
\IEEEcompsocthanksitem Jianhua Tao is with the Department of Automation, Tsinghua University, Beijing, China, 100084.\protect\\
E-mail: jhtao@tsinghua.edu.cn
}
\thanks{Manuscript received xxx; revised xxx. (Corresponding authors: Bin Liu and Jianhua Tao)}}

%
%

\markboth{Journal of \LaTeX\ Class Files,~Vol.~14, No.~8, August~2015}%
{Shell \MakeLowercase{\textit{et al.}}: Bare Demo of IEEEtran.cls for Computer Society Journals}
%



\IEEEtitleabstractindextext{%
\begin{abstract}
With the proliferation of user-generated online videos, Multimodal Sentiment Analysis (MSA) has attracted increasing attention recently. Despite significant progress, there are still two major challenges on the way towards robust MSA: 1) inefficiency when modeling cross-modal interactions in unaligned multimodal data; and 2) vulnerability to random modality feature missing which typically occurs in realistic settings. In this paper, we propose a generic and unified framework to address them, named Efficient Multimodal Transformer with Dual-Level Feature Restoration (EMT-DLFR). Concretely, EMT employs utterance-level representations from each modality as the global multimodal context to interact with local unimodal features and mutually promote each other. It not only avoids the quadratic scaling cost of previous local-local cross-modal interaction methods but also leads to better performance. To improve model robustness in the incomplete modality setting, on the one hand, DLFR performs low-level feature reconstruction to implicitly encourage the model to learn semantic information from incomplete data. On the other hand, it innovatively regards complete and incomplete data as two different views of one sample and utilizes siamese representation learning to explicitly attract their high-level representations. Comprehensive experiments on three popular datasets demonstrate that our method achieves superior performance in both complete and incomplete modality settings.
\end{abstract}

}

\maketitle

\IEEEdisplaynontitleabstractindextext

%
\IEEEpeerreviewmaketitle

\IEEEraisesectionheading{\section{Introduction}\label{sec:introduction}}

%
%
%
%
\IEEEPARstart{M}{ultimodal} Sentiment Analysis (MSA), which leverages multimodal signals to achieve an affective understanding of user-generated video, has become an active research area due to its wide applications in marketing management \cite{soleymani2017survey, poria2020beneath}, social media analysis \cite{somandepalli2021computational, stappen2021multimodal}, and human-computer interaction \cite{cambria2017affective, poria2017review}, to name a few.
It mainly involves sequential data of three common modalities, i.e., audio (acoustic behaviors), vision (facial expressions), and text (spoken words). These different types of data provide us with abundant information to make a thorough understanding of human sentiment.
Nevertheless, it remains challenging to efficiently fuse the heterogeneous sequential features in practical applications. 
The first issue is that multimodal sequences usually exhibit \textit{unaligned} nature as different modalities typically have variable sampling rates. 
In addition, they often suffer from random modality feature missing (i.e., \textit{incomplete}) due to many inevitable factors in real-world scenarios. For instance, the speech may be temporarily corrupted by background noise or sensor failure. The speaker's face could occasionally miss because of occlusion and motion. Some spoken words are probably unavailable owing to automatic speech recognition errors.

The earlier studies address the \textit{unaligned} issue by manually performing forced word-level alignment before model training \cite{poria2017context, zadeh2018memory, zadeh2018multi, gu2018multimodal, pham2019found}. However, the manual alignment process requires domain expert knowledge and is not always feasible in the real-world deployment of MSA models. 
Recently, Multimodal Transformer (MulT) \cite{tsai2019multimodal} has been proposed to directly model cross-modal correlations in unaligned multimodal sequences. It utilizes directional pairwise cross-modal attention to attend to dense (i.e., \textit{local-local}) interactions across distinct time steps between two involved modalities. 
Although MulT can address the unaligned issue, it is not efficient to conduct multimodal fusion in a pairwise manner. Therefore, Lv et al. \cite{lv2021progressive} propose the Progressive Modality Reinforcement (PMR), which introduces a message hub to communicate with each modality. 
The message hub can send common messages to each modality and it can also collect information from them.
In this way, PMR avoids the inefficient pairwise communication of two modalities in MulT.
Unfortunately, both PMR and MulT suffer from quadratic computational complexity over the involved modalities, as they focus on modeling the \textit{local-local} cross-modal dependencies across all modalities (see details in Section \ref{sec_method_emt}).

Moreover, random modality feature missing which often occurs in realistic settings exacerbates the problem of fusing unaligned multimodal sequences. 
A simple method to tackle the \textit{incomplete} issue is to perform zero, average, or nearest-neighbor imputation during model inference. However, there is a large domain gap between the ground truth and the imputed one. Thus, the model performance typically degrades severely using these naive imputations.
Recently, several modality translation-based methods \cite{pham2019found, zhao2021missing, tang2021ctfn} have been proposed to learn robust joint multimodal representations that retain maximal information from all modalities. While appealing, these methods are developed to handle the entire loss of one or more modalities which is less likely to happen in practice. Besides, most of them are only applicable to aligned multimodal inputs.
More recently, Yuan et al. \cite{yuan2021transformer} introduce the Transformer-based Feature Reconstruction Network (TFR-Net) to cope with random modality feature missing. It expects the model to implicitly learn semantic features from incomplete multimodal sequences by reconstructing the missing parts. Though achieving promising results, TFR-Net has the risk of learning trivial solutions as the model may find a shortcut instead of inferring semantics to solve the reconstruction task \cite{jing2020self}. Besides, it builds on MulT and thus also has a quadratic scaling cost with the involved modalities.

To efficiently fuse unaligned multimodal sequences in both complete and incomplete modality settings, we propose a generic and unified framework in this paper, named Efficient Multimodal Transformer with Dual-Level Feature Restoration (EMT-DLFR). 
In contrast to MulT and PMR, the Efficient Multimodal Transformer (EMT) explores less dense (i.e., \textit{global-local}) cross-modal interactions, which is mainly motivated by the observation that a large amount of redundancy exists in unaligned multimodal sequences (especially for the audio and video modalities which have high sampling rates). 
Inspired by the bottleneck mechanism in Transformers \cite{lee2019set,jaegle2021perceiver,nagrani2021attention}, EMT utilizes utterance-level representations from each modality as the \textit{global} multimodal context to interact with \textit{local} unimodal features. 
On the one side, local unimodal features can be efficiently reinforced by the global multimodal context via cross-modal attention. In turn, the global multimodal context can update itself by extracting useful information from local unimodal features through symmetric cross-modal attention. 
By stacking multiple layers, the global multimodal context and local unimodal features can mutually promote each other and refine themselves progressively. 
Thanks to the introduction of the global multimodal context, EMT not only practically has linear computational complexity over the involved modalities but also leads to performance gains. 
Furthermore, we introduce hierarchical parameter sharing for EMT to increase parameter efficiency and ease model training.

To promote model robustness to random modality feature missing, we randomly mask the input feature sequences of complete data to mimic real-world scenarios and utilize the Dual-Level Feature Restoration (DLFR) built upon EMT to achieve robust representation learning from incomplete multimodal data. On the one hand, DLFR follows TFR-Net and tries to reconstruct the missing parts by exploiting available intra- and inter-modal clues using multiple stacked layers in EMT, i.e, the Low-Level Feature Restoration (LLFR), or more specifically, \textit{low-level} feature reconstruction. In this way, LLFR \textit{implicitly} encourages the model to learn semantic information from incomplete multimodal sequences. On the other hand, inspired by recent advances in self-supervised representation learning \cite{he2020momentum, chen2020simple, chen2021exploring}, we originally regard incomplete and complete sequences as two different views of one sample, and utilize siamese representation learning to \textit{explicitly} attract high-level representations of incomplete and complete views in the latent space, i.e., High-Level Feature Restoration (HLFR), or more specifically, \textit{high-level} feature attraction. Compare with LLFR, HLFR is more direct and thus more effective. Nevertheless, they are complementary to each other. Therefore, the combination of LLFR and HLFR can be a unified framework for robust MSA. 

To verify the effectiveness of the proposed method, we conduct comprehensive experiments on three widely used MSA benchmark datasets, including CMU-MOSI \cite{zadeh2016multimodal}, CMU-MOSEI \cite{zadeh2018multimodal}, and CH-SIMS \cite{yu2020ch}. 
The results show that our proposed method outperforms previous state-of-the-art methods in both incomplete and complete modality settings.
To summarize, the main contributions of this paper are as follows:
\begin{itemize}
\item We propose EMT, an Efficient Multimodal Transformer to achieve effective and efficient fusion of unaligned multimodal data. It not only avoids the quadratic scaling cost of previous dense cross-modal interaction methods but also achieves better performance than them. 
\item We propose DLFR, a Dual-Level Feature Restoration method to improve model robustness to random modality feature missing which typically occurs in real-world scenarios. It combines both implicit low-level feature reconstruction and explicit high-level feature attraction to realize robust representation learning from incomplete multimodal data. 
\item Extensive experiments on three popular MSA benchmark datasets demonstrate that EMT and EMT-DLFR achieve state-of-the-art performance in the complete and incomplete modality settings, respectively. \footnote{The code will be available at \url{https://github.com/sunlicai/EMT-DLFR}.}
\end{itemize}

\section{Related Work}

\subsection{MSA in the Complete Modality Setting}
Most works in MSA assume a complete modality setting and they are centered around developing various methods to fuse heterogeneous features from different modalities.
Generally, they conduct multimodal fusion at two different levels: 1) utterance level, and 2) element level. The utterance-level fusion methods mainly include simple concatenation \cite{morency2011towards, perez2013utterance, yu2021learning, han2021improving}, attention \cite{poria2017context, lian2019conversational, hazarika2020misa}, and tensor-based fusion \cite{zadeh2017tensor, liu2018efficient, mai2019divide, jin2020dual}. Although these methods obtain promising results, they ignore the fine-grained cross-modal interactions. Therefore, lots of methods have been proposed to perform element-level fusion on manually aligned multimodal sequences, including recurrent methods \cite{rajagopalan2016extending, zadeh2018memory, zadeh2018multi}, attention-based methods \cite{chen2017multimodal, gu2018multimodal}, and multimodal-aware word embeddings \cite{wang2019words, rahman2020integrating}.
Unfortunately, the manual alignment process requires domain expert knowledge and is not always feasible in real-world applications. Motivated by the great success of Transformer \cite{vaswani2017attention} in various fields of deep learning, Transformer-based methods \cite{tsai2019multimodal, zadeh2019factorized, lv2021progressive, liang2021attention, sun2021multimodal, lian2021ctnet, rajan2022cross, yuan2021transformer} have attracted increasing attention in recent years as they can directly perform multimodal fusion on unaligned multimodal sequences. For instance, MulT \cite{tsai2019multimodal} utilizes directional pairwise cross-modal attention to attend to dense interactions between different modalities. PMR \cite{lv2021progressive} further introduces a message hub to explore multi-way interactions in a single cross-modal attention module.
TFR-Net \cite{yuan2021transformer} adds an extra intra-modal Transformer in parallel with the cross-modal Transformer in MulT to achieve simultaneous modeling of intra-modal and cross-modal interactions.
Although achieving encouraging results, they all have quadratic computational complexity over the involved modalities. In contrast, due to the introduction of the global multimodal context, our proposed method enjoys the linear scaling cost in practice and even has better performance.

\subsection{MSA in the Incomplete Modality Setting}
Compared with numerous methods in the above setting, there are only a few studies focusing on improving model robustness in the incomplete setting, despite its critical role in enabling reliable deployment in the wild.
Parthasarathy et al. \cite{parthasarathy2020training} propose a strategy to cope with incomplete sequences by randomly ablating input features during training. Similarly, Hazarika et al. \cite{hazarika2022analyzing} utilize modality-perturbation (i.e., removing or masking modalities) training to reduce the sensitivity of the model to the missing language modality. 
Liang et al. \cite{liang2019learning} present a tensor rank regularization method to learn multimodal representations from aligned imperfect time series data.
Pham et al. \cite{pham2019found} utilize sequential cyclic translations to learn robust joint representations that may not require all modalities as input at inference time.
Tang et al. \cite{tang2021ctfn} further explore pairwise bidirectional modality translations.
Moreover, Zhao et al. \cite{zhao2021missing} employ data augmentation and cyclic translations based on cascaded residual autoencoder \cite{tran2017missing} to tackle the uncertain missing modality issue.
However, these modality translation-based methods are developed to handle the entire loss of one or more modalities which is less likely to happen in real-world scenarios. Besides, most of them only accept aligned sequences as inputs.
Recently, Yuan et al. \cite{yuan2021transformer} propose to reconstruct low-level missing features to implicitly force the model to learn semantic features from incomplete multimodal sequences. 
Lian et al. \cite{lian2023gcnet} also introduce low-level feature reconstruction for incomplete multimodal learning in conversations. 
Compared with these two methods, our proposed method additionally utilizes explicit high-level feature attraction to improve model robustness. Moreover, our experiments demonstrate that explicit high-level feature attraction is superior to implicit low-level feature reconstruction and they work best when combined with each other.

\section{Method}
\begin{figure*}[t]
	\centering
	\includegraphics[height=0.5\linewidth, width=1.0\linewidth]{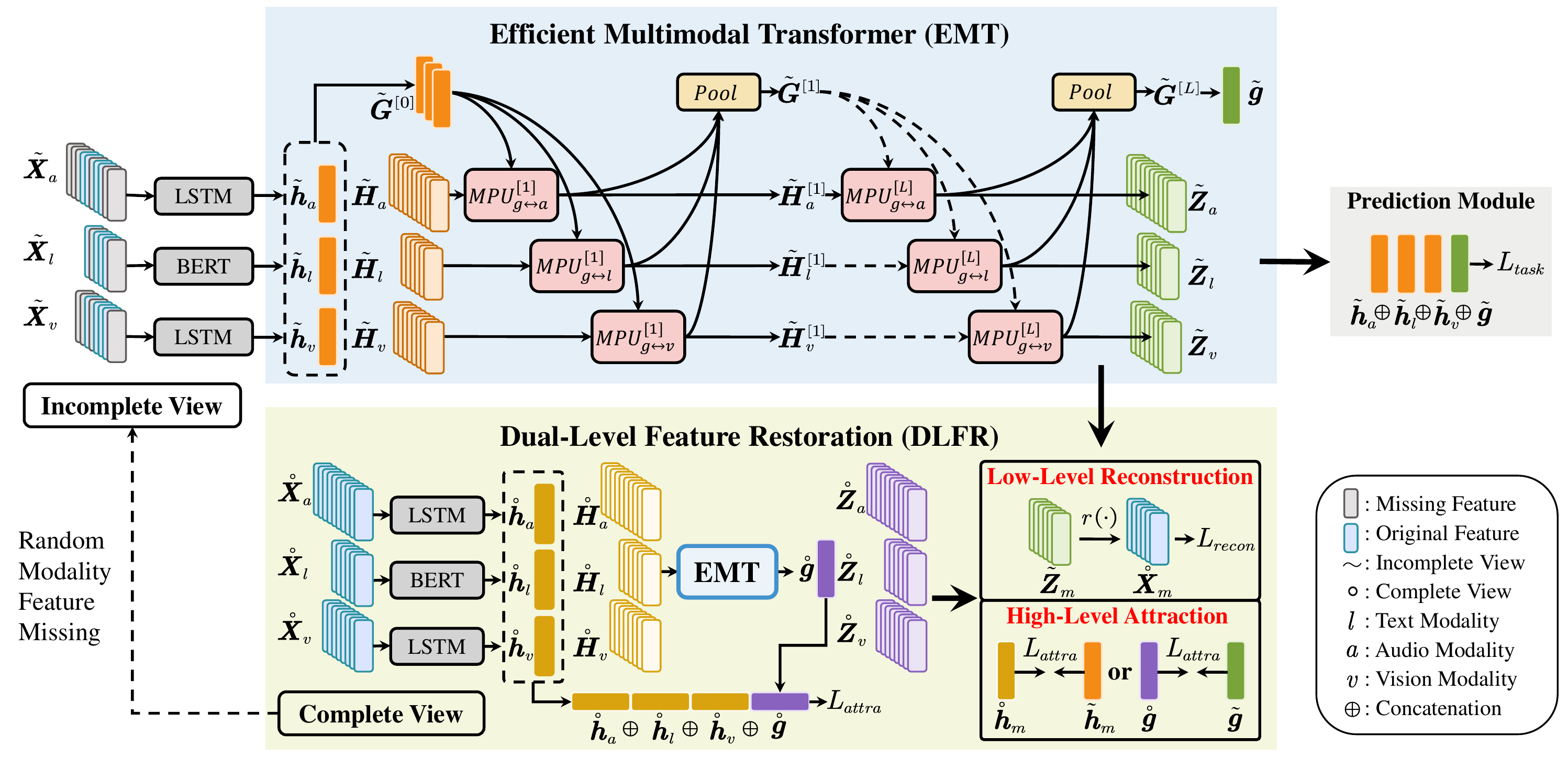}
	\caption{The overall architecture of EMT-DLFR, which mainly consists of the Efficient Multimodal Transformer (EMT) and Dual-Level Feature Restoration (DLFR) for efficient and robust MSA. To achieve efficient multimodal fusion, EMT utilizes utterance-level representations from each modality as the global multimodal context $G$ to interact with local unimodal features $H_m$ ($m \in \{l,a,v\}$) through the Mutual Promotion Unit (MPU). Based on EMT, DLFR employs both explicit high-level feature attraction and implicit low-level feature reconstruction to guide robust representation learning from incomplete multimodal sequences. Best viewed in color.}
	\label{fig_emt_dlfr}
\end{figure*}

In this section, we describe the proposed Efficient Multimodal Transformer with Dual-Level Feature Restoration (EMT-DLFR) for robust MSA, with its whole pipeline in the incomplete modality setting shown in Fig. \ref{fig_emt_dlfr}. 
In this setting, we use the ground-truth complete view to achieve robust representation learning from incomplete multimodal sequences. 
Specifically, for both views, we first utilize modality-specific encoders to obtain utterance-level (i.e., global) and element-level (i.e., local) intra-modal features from different modality inputs. 
Then the Efficient Multimodal Transformer (EMT) is employed to effectively and efficiently capture useful cross-modal interactions between the global multimodal context and local unimodal features. After that, utterance-level intra- and inter-modal features are combined to get the final sentiment intensity prediction.
To promote model robustness to random modality feature missing, the Dual-Level Feature Restoration (DLFR) conducts high-level feature attraction and low-level feature reconstruction simultaneously. The former explicitly attracts high-level intra- and inter-modal representations of two views in the latent space, while the latter reconstructs low-level modality inputs from complete view to implicitly urge the model to learn semantic information.
In the complete modality setting, the problem degenerates into a simple case. Thus, we do not perform DLFR and only the original prediction loss is used in this setting.

In the following parts, we begin by giving a problem definition. 
Then we elaborate on the four main modules in EMT-DLFR: unimodal feature encoder (Section \ref{sec_method_enc}), EMT (Section \ref{sec_method_emt}), prediction module (Section \ref{sec_method_pred}), and DLFR (Section \ref{sec_method_dlfr}). Finally, we present the overall loss function for model training in two modality settings (Section \ref{sec_method_loss}).

\subsection{Problem Definition}
The task of MSA is to predict the sentiment intensity of the speaker in the video. Typically, three modalities are involved in MSA, i.e., text or language ($l$), audio ($a$), and vision ($v$). We denote the complete input feature sequences as $\mathring{X}_m \in \mathbb{R}^{T_m \times f_m}$, where $T_m$ is sequence length and $f_m$ is the feature dimension of modality $m \in \{l, a, v\}$. Specifically,  $\mathring{X}_a$ and $\mathring{X}_v$ are shallow features extracted by open-source tools, while $\mathring{X}_l$ is the raw text tokens output by the BERT \cite{devlin2019bert} tokenizer.
To mimic random modality feature missing in real-world scenarios, we randomly mask the complete sequence $\mathring{X}_m$ to obtain the incomplete sequence $\tilde{X}_m= F(\mathring{X}_m, g_m) \in \mathbb{R}^{T_m \times f_m}$, where $F(\cdot)$ is the mask function,  $g_m \in \{0,1\}^{T_m}$ is a random temporal mask which indicates the positions to mask. 
For the audio and video modality, the mask function replaces the original feature vector in the masked position with a zero vector. For the text modality, it replaces the original token with the [UNK] token in BERT vocabulary \cite{yuan2021transformer}.
Our goal is to develop a robust model which can efficiently integrate all the available multimodal information to make an accurate prediction of sentiment intensity score $y \in \mathbb{R}$ in both complete and incomplete modality settings.
Note that, when we do not use the diacritical mark $\mathring{}$ or $\tilde{}$ (e.g., $X$), it indicates that both views can be applied.

\subsection{Unimodal Feature Encoder}
\label{sec_method_enc}
Following previous works \cite{hazarika2020misa,yuan2021transformer, yu2021learning}, we use the powerful BERT \cite{devlin2019bert} model to encode raw text tokens into contextual word embeddings. For the audio and video modality, we opt for simplicity and employ the commonly used Long Short-Term Memory (LSTM) recurrent neural network \cite{hochreiter1997long} to capture the temporal dependencies in the feature sequence, i.e.,
\begin{equation}
\begin{split}
H_l &= \textrm{BERT}(X_l) \\
H_a &= \textrm{LSTM}(X_a) \\
H_v &= \textrm{LSTM}(X_v) 
\end{split}
\label{eq_intra}
\end{equation}
where $H_m \in \mathbb{R}^{T_m \times d_m}, m \in \{l,a,v\}$. Since the [CLS] token in BERT aggregates the information from all tokens, we use its embedding as the utterance-level representation of the text modality $h_l \in \mathbb{R}^{d_l}$. For $h_a \in \mathbb{R}^{d_a}$ and $h_v \in \mathbb{R}^{d_v}$, we simply use the feature of the last time step in $H_a$ and $H_v$. 
Finally, we use several linear layers to project $H_l$, $H_a$, $H_v$, $h_l$, $h_a$, and $h_v$ to the same dimension $d$ to facilitate subsequent multimodal fusion.

\subsection{Efficient Multimodal Transformer}
\label{sec_method_emt}
In this part, we first give a preliminary introduction. 
Then we introduce the basic building block of EMT, named Mutual Promotion Unit (MPU). 
Based on MPU, we elaborate on three multimodal fusion strategies for modeling cross-modal interactions.
The first two are inspired by two predominant models (i.e., MulT \cite{tsai2019multimodal} and PMR \cite{lv2021progressive}) in MSA. 
However, both of them suffer from quadratic computational complexity over the involved modalities. The third practically has a linear scaling cost and thus is adopted in EMT by default. 
Finally, we present hierarchical parameter sharing for EMT to further improve parameter efficiency.

\subsubsection{Preliminary}
 Self-attention (SA) is the core component in Transformer. It allows modeling of global dependencies in a sequence via scaled dot-product attention \cite{vaswani2017attention}. For the input sequence $H_t \in \mathbb{R}^{T_t \times d}$, we define the Querys as $Q_t = H_t W_Q$, Keys as $K_t = H_t W_K$, Values as $V_t = H_t W_V$, where $W_Q$ and $W_K \in \mathbb{R}^{d \times d_k}$, $W_V$ $ \in \mathbb{R}^{d \times d_v}$. Then SA can be formulated as follows:
\begin{equation}
\textrm{SA}(H_t) = \textrm{softmax}(\frac{Q_t K_t^T}{\sqrt{d_k}})V_t
\label{eq_sa}
\end{equation}

Cross-modal attention (CA) involves two modalities. The Querys are from the target modality $t$, while the Keys and Values are from the source modality $s$, i.e., $Q_t = H_t W_Q$, $K_s = H_s W_K$, $V_s = H_s W_V$. In this way, CA can provide a latent adaptation from modality $s$ to $t$:
\begin{equation}
\textrm{CA}(H_t, H_s) = \textrm{softmax}(\frac{Q_t K_s^T}{\sqrt{d_k}})V_s
\label{eq_ca}
\end{equation}
which is a good way to fuse cross-modal information \cite{tsai2019multimodal}.

Note that, for simplicity, we only present the formulation of single-head attention. In practice, we use multi-head SA or CA (i.e., MHSA or MHCA) to allow the model to attend to information from different feature subspaces \cite{vaswani2017attention}.

\subsubsection{Mutual Promotion Unit}

\begin{figure}[t]
	\centering
	\includegraphics[width=0.9\linewidth]{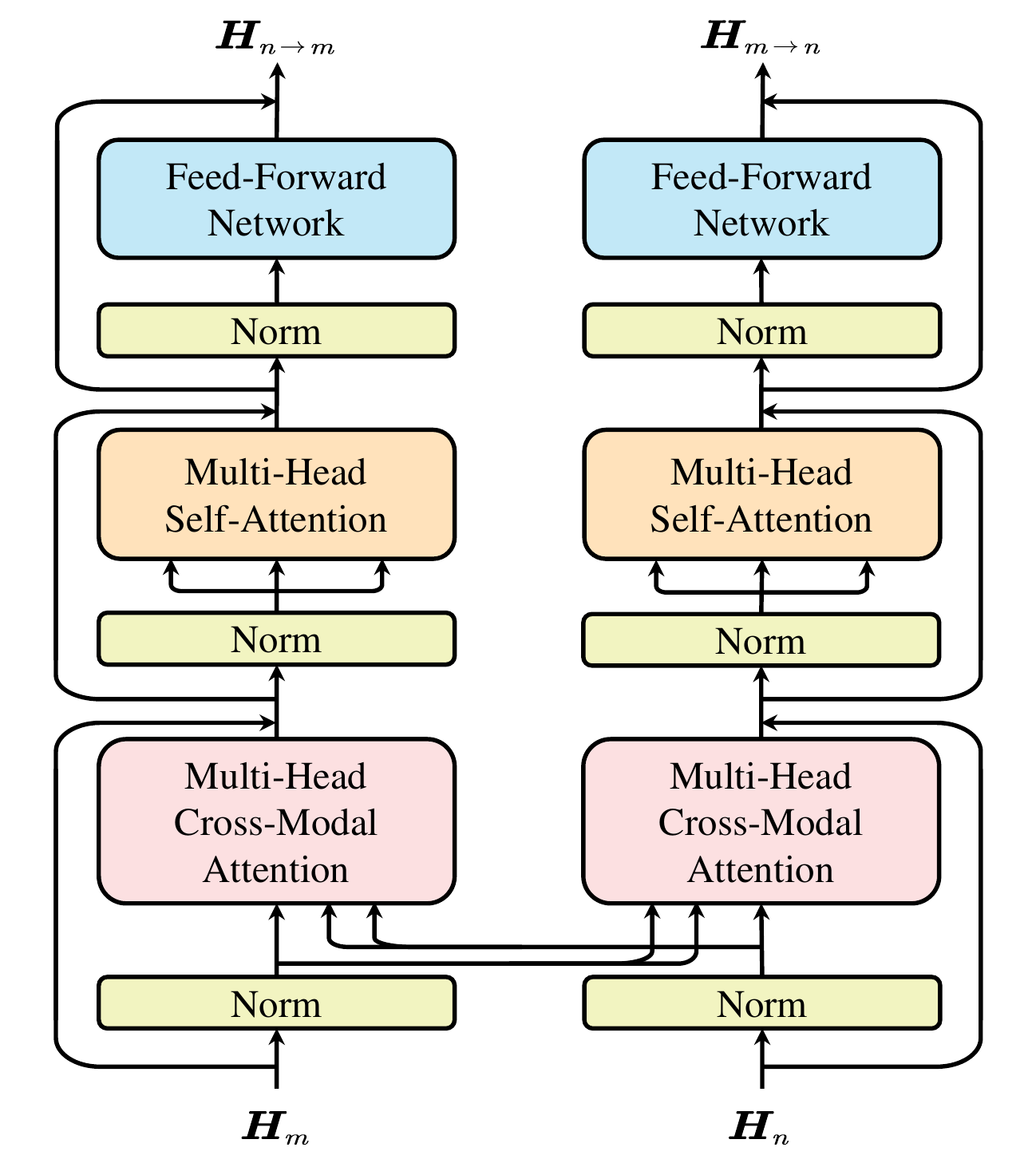}
	\caption{The architecture of Mutual Promotion Unit (MPU).}
	\label{fig_mpu}
\end{figure}

The architecture of MPU is shown in Fig. \ref{fig_mpu}. 
It mainly utilizes symmetric cross-modal attention to explore inherent correlations between elements across two input feature sequences. 
In this way, MPU enables useful information exchange between two sequences and thus they can \textit{mutually promote} each other. 
To allow further information integration, self-attention is employed to model temporal dependencies in each feature sequence. 
Formally, MPU takes two sequences $H_m$ and $H_n$ as inputs and outputs their mutually promoted ones $H_{m \rightarrow n}$ and $H_{n \rightarrow m}$:
\begin{equation}
\begin{split}
H_{n \rightarrow m}, H_{m \rightarrow n} &= \textrm{MPU}_{m \leftrightarrow n}(H_m, H_n) \\
H_{n \rightarrow m} &= \textrm{MPU}_{n \rightarrow m}(H_m, H_n) \\
H_{m \rightarrow n} &= \textrm{MPU}_{m \rightarrow n}(H_n, H_m) \\
\end{split}
\label{eq_mpu}
\end{equation}
where $\rightarrow$ and $\leftrightarrow$ indicates the direction of information flow.
In specific, the calculation of $MPU_{n \rightarrow m}(H_m, H_n)$ is as follows:
\begin{equation}
\begin{split}
H_{n \rightarrow m}^{'}  &= \textrm{MHCA}(\textrm{LN}(H_m), \textrm{LN}(H_n)) + H_m, \\
H_{n \rightarrow m}^{''}  &= \textrm{MHSA}(\textrm{LN}(H_{n \rightarrow m}^{'})) + H_{n \rightarrow m}^{'}, \\
H_{n \rightarrow m}  &= \textrm{FFN}(\textrm{LN}(H_{n \rightarrow m}^{''})) + H_{n \rightarrow m}^{''}
\end{split}
\label{eq_mpu_one_side}
\end{equation}
where $\textrm{LN}$ denotes layer normalization \cite{ba2016layer}, $\textrm{FFN}$ is the position-wise feed-forward network in Transformer. 
$\textrm{MPU}_{m \rightarrow n}(H_n, H_m)$ can be formulated in a similar way.

Considering that the computational time complexity of $\textrm{MHCA}(H_m, H_n)$ is $O(T_m T_n)$ and $\textrm{MHSA}(H_{m})$ has the complexity of $O(T_m^2)$, the total time complexity of an MPU is $O(T_m T_n + T_m^2 + T_n T_m + T_n^2) = O((T_m + T_n)^2)$.

\subsubsection{Multimodal Fusion Strategy}
Built upon MPU, we now describe three multimodal fusion strategies in a progressive manner and give a computational complexity analysis for each of them (summarized in Table \ref{tab_fusion_complexity}). Note that the third fusion strategy (Fig. \ref{fig_emt_dlfr}) is adopted in EMT by default and we leave their empirical comparisons in the ablation study (Section \ref{sec_ablation}).
For simplicity, we only present the information flow from layer $i$ to $i+1$ in an EMT with $L$ layers.
We suppose that the input feature sequence is $H_m^{[0]} = H_m + \textrm{PE}_m \in \mathbb{R}^{T_m \times d}$ and the corresponding utterance-level representation is $h_m^{[0]} = h_m \in \mathbb{R}^{d}$, where $m \in \{l, a, v\}$ and $\textrm{PE}_m \in \mathbb{R}^{T_m \times d}$ is the sinusoidal position embedding in the vanilla Transformer.

\begin{table}[]
\centering
\caption{Computational complexity analysis of three multimodal fusion strategies. $M$ is the number of involved modalities, $T$ is the temporal length of the input feature sequence. For simplicity, we assume aligned multimodal inputs. The detailed time complexity for unaligned inputs can refer to the main text.}
\label{tab_fusion_complexity}
\resizebox{\linewidth}{!}{
\begin{tabular}{ccc}
\toprule
Fusion Strategy     & Space Complexity         & Time Complexity   \\
\midrule
OOLL     & $O(M^2)$  & $O(M^2T^2)$    \\
OALL     & $O(M)$    & $O(M^2T^2)$  \\
OAGL     & $O(M)$    & $O(MT^2)$     \\
\bottomrule
\end{tabular}
}
\end{table}

\begin{itemize}
\item \textbf{One-to-One Local-Local Fusion.}
This fusion strategy shares the same spirit with MulT \cite{tsai2019multimodal}. 
The core is to explore \textit{one-to-one local-local} (OOLL) cross-modal interactions across all modalities by applying MPU to each paired modalities, i.e.,
\begin{equation}
\begin{split}
H_{a \rightarrow l}^{[i]}, H_{l \rightarrow a}^{[i]} &= \textrm{MPU}_{l \leftrightarrow a}^{[i]}(H_l^{[i]}, H_a^{[i]}) \\
H_{v \rightarrow l}^{[i]}, H_{l \rightarrow v}^{[i]} &= \textrm{MPU}_{l \leftrightarrow v}^{[i]}(H_l^{[i]}, H_v^{[i]}) \\
H_{v \rightarrow a}^{[i]}, H_{a \rightarrow v}^{[i]} &= \textrm{MPU}_{a \leftrightarrow v}^{[i]}(H_a^{[i]}, H_v^{[i]}) \\
\end{split}
\label{eq_ooll}
\end{equation}
When there are $M$ modalities involved in the fusion process, the number of required MPUs in each layer is $C(M,2)=\frac{1}{2}M(M-1)$, and the time complexity of OOLL is $O(\frac{1}{2}\sum_{m=1}^{M}{(\sum_{n \neq m}{(T_m + T_n)^2})})$. If all modalities are aligned (i.e., $T_m=T, \forall m$), the time complexity degenerates into $O(M^2T^2)$. 
Thus, OOLL suffers from quadratic space and time complexity over the involved modalities.

\item \textbf{One-to-All Local-Local Fusion.}
Since it is inefficient to fuse multiple modalities in a one-to-one manner, this strategy utilizes the common message introduced in PMR \cite{lv2021progressive} to explore \textit{one-to-all local-local} (OALL) cross-modal interactions in a single MPU. 
The common message $C$ is initialized by temporal concatenation of unimodal feature sequences, 
i.e., $C^{[0]} = \textrm{Concat}(H_l^{[0]}, H_a^{[0]}, H_v^{[0]}) \in \mathbb{R}^{(T_l+T_a+T_v) \times d}$. Formally, we have
\begin{equation}
\begin{split}
H^{[i+1]}_l, C_{l \rightarrow c}^{[i]} &= \textrm{MPU}_{l \leftrightarrow c}^{[i]}(H_l^{[i]}, C^{[i]}) \\
H^{[i+1]}_a, C_{a \rightarrow c}^{[i]} &= \textrm{MPU}_{a \leftrightarrow c}^{[i]}(H_a^{[i]}, C^{[i]}) \\
H^{[i+1]}_v, C_{v \rightarrow c}^{[i]} &= \textrm{MPU}_{v \leftrightarrow c}^{[i]}(H_v^{[i]}, C^{[i]}) \\
\end{split}
\label{eq_oall}
\end{equation}
Due to the introduction of the common message, OALL only requires $M$ MPUs in each layer.
After conducting a similar complexity analysis, we can get the time complexity of OALL: $O(\sum_{m=1}^{M}{(T_m + \sum_{n=1}^{M}{(T_n)})^2})$, which is greater than that of OOLL. In the modality-aligned setting, it degenerates into $O(M^2T^2)$, which indicates that this strategy also has a quadratic scaling cost.

\item \textbf{One-to-All Global-Local Fusion.}
The above analysis indicates that both OOLL and OALL have quadratic computational complexity over the involved modalities. We observe that this issue results from the fact that they both explore local-local cross-modal interactions across all modalities. 
We believe that it is neither efficient nor necessary to perform multimodal fusion in a local-local fashion.
On the one hand, a large amount of redundancy exists in the unimodal feature sequence, especially for the audio and video modality which has high sampling rates. 
On the other hand, too many local features in one modality (or the common message) may distract the attention of another modality during their interactions because the latter could not `see' the global (i.e., summarized) information of the former.
Moreover, it could increase the risk of overfitting spurious cross-modal correlation signals. 
Therefore, we believe that the utterance-level representations from each modality can substitute for the common message in OALL and serves as the \textit{global} multimodal context $G$ to interact with \textit{local} unimodal features, i.e., 
\begin{equation}
\begin{split}
H^{[i+1]}_l, G_{l \rightarrow g}^{[i]} &= \textrm{MPU}_{l \leftrightarrow g}^{[i]}(H_l^{[i]}, G^{[i]}) \\
H^{[i+1]}_a, G_{a \rightarrow g}^{[i]} &= \textrm{MPU}_{a \leftrightarrow g}^{[i]}(H_a^{[i]}, G^{[i]}) \\
H^{[i+1]}_v, G_{v \rightarrow g}^{[i]} &= \textrm{MPU}_{v \leftrightarrow g}^{[i]}(H_v^{[i]}, G^{[i]}) \\
\end{split}
\label{eq_oagl}
\end{equation}
where $G^{[0]}=\textrm{Concat}(h_l^{[0]}, h_a^{[0]}, h_v^{[0]}) \in \mathbb{R}^{3 \times d}$.
In the above way, this strategy can capture \textit{one-to-all global-local} (OAGL) cross-modal interactions in a single MPU.
By stacking multiple layers, the global multimodal context and local unimodal features can mutually promote each other and refine themselves progressively. 
Same as OALL, OAGL requires $M$ MPUs in each layer. However, thanks to the global multimodal context, the time complexity of OAGL decreases to $O(\sum_{m=1}^{M}{(T_m + M)^2}) \approx O(\sum_{m=1}^{M}{T_m^2})$ (practically, we have $M \ll T_m$), and it degenerates into $O(MT^2)$ in the modality-aligned setting. 
Thus, the default OAGL fusion strategy in EMT not only has linear space complexity but also enjoys linear computations over the involved modalities.

\end{itemize}

Finally, we briefly introduce the pooling layer (Fig. \ref{fig_emt_dlfr}) for aggregating promoted information from different modalities to facilitate subsequent fusion.
Specifically, we utilize an attention-based pooling layer to implement it. 
Taking OAGL as an example, we have three promoted global multimodal contexts, i.e., $G_{l \rightarrow g}^{[i]}$, $G_{a \rightarrow g}^{[i]}$, and $G_{v \rightarrow g}^{[i]}$. The new global multimodal context can be obtained as follows:
\begin{equation}
G^{[i+1]}=\textrm{softmax}(v^T\tanh{(W^T G_{g}^T + b))} G_{g}
\label{eq_pool}
\end{equation}
where $G_{g}=\textrm{Concat}(G_{l \rightarrow g}^{[i]}, G_{a \rightarrow g}^{[i]}, $ $G_{v \rightarrow g}^{[i]})$,
$v$, $W$, and $b$ are learnable parameters. 
We also develop other two pooling methods, including average pooling and MLP-based pooling. 
However, they have inferior performance (Section \ref{sec_ablation}).

\subsubsection{Hierarchical Parameter Sharing}
Inspired by \cite{lan2019albert,jaegle2021perceiver}, we propose hierarchical parameter sharing to further improve model parameter efficiency.
Concretely, there are three levels to share parameters in EMT, including MPU-level sharing, modality-level sharing, and layer-level sharing. 
The first level shares the parameters of two directional sub-MPUs in an MPU, i.e., $\textrm{MPU}_{n \rightarrow m}$ and $\textrm{MPU}_{m \rightarrow n}$ in Equation \ref{eq_mpu}.
The second level shares MPUs across modalities, i.e., $\textrm{MPU}_{m \leftrightarrow g}^{[i]}$ ($m\in \{l,a,v\}$) in Equation \ref{eq_oagl}. 
And the last one shares parameters across layers.
We show that the hierarchical parameter sharing strategy not only improves parameter efficiency but also can make the optimization of EMT easier in Section \ref{sec_ablation}.

\subsection{Prediction Module}
\label{sec_method_pred}
After multimodal fusion, we flatten the temporal dimension of the final global multimodal context (i.e., $G^{[L]}$) in EMT to get utterance-level inter-modal representation $g \in \mathbb{R}^{3d}$. We then concatenate $g$ and utterance-level intra-modal representations $\{h_m\}$ ($m \in \{l, a, v\}$) and finally pass them through a multi-layer perceptron (MLP) to get the sentiment prediction $y'$. 
Following previous works \cite{yu2021learning,yuan2021transformer}, we employ L1 loss as the prediction loss, i.e.,
\begin{equation}
\mathcal{L}_{\textrm{task}} = |y - y'| 
\label{eq_loss_pred}
\end{equation}

\subsection{Dual-Level Feature Restoration}
\label{sec_method_dlfr}
To improve model robustness in the incomplete modality setting, we utilize the complete view to perform the dual-level feature restoration (DLFR), including \textit{implicit} low-level feature reconstruction and \textit{explicit} high-level feature attraction. Note that, the complete view is only used to guide representation learning during training, and it will not be available at the test stage.

\subsubsection{Low-Level Feature Reconstruction}
Reconstruction-based training objective (e.g., masked autoencoding) has achieved great success for representation learning in the fields of computer vision \cite{vincent2008extracting, pathak2016context, he2022masked}, natural language processing \cite{devlin2019bert, radford2018improving}, and speech signal processing \cite{baevski2020wav2vec, hsu2021hubert}. Motivated by this, we leverage low-level feature restoration (LLFR), i.e., low-level feature reconstruction, to \textit{implicitly} encourage the model to learn semantic representations from incomplete multimodal inputs.

After multimodal fusion via EMT, the incomplete sequence $\tilde{X}_m$ is mapped to a latent sequence $\tilde{Z}_m=\tilde{H}_m^{[L]}$ that has sufficient awareness of both global inter-modal and local intra-modal information. Thus, we pass $\tilde{Z}_m$ through a simple MLP-based decoder $r(\cdot)$ to reconstruct the complete sequence $\mathring{X}_m$. Note that, since reconstructing raw text tokens will waste a large amount of model capacity\footnote{Reconstructing raw text tokens from text representations will need a very big projection matrix $W \in \mathbb{R}^{30522 \times 768}$ with about 23.4M parameters (30522 is the vocabulary size of BERT, 768 is the dimension of text representations).}, we use the output embedding of the BERT model $\mathring{H}_l$ (instead of raw text token sequence $\mathring{X}_l$) as the reconstruction target of the text modality. 
Following \cite{yuan2021transformer}, we employ the smooth L1 loss to evaluate the reconstruction quality, i.e.,
\begin{equation}
\begin{split}
\mathcal{L}^l_{\textrm{recon}} &= \textrm{smooth}_{\textrm{L1}}((\mathring{H}_l - r(\tilde{Z}_l)) \cdot (1-g_l)) \\
\mathcal{L}^a_{\textrm{recon}} &= \textrm{smooth}_{\textrm{L1}}((\mathring{X}_a - r(\tilde{Z}_a)) \cdot (1-g_a)) \\
\mathcal{L}^v_{\textrm{recon}} &= \textrm{smooth}_{\textrm{L1}}((\mathring{X}_v - r(\tilde{Z}_v)) \cdot (1-g_v)) \\
\mathcal{L}_{\textrm{recon}}   &= \sum_{m \in \{l,a,v\}}\mathcal{L}^m_{\textrm{recon}}
\label{eq_loss_llfr}
\end{split}
\end{equation}
where, 
\begin{equation}
\textrm{smooth}_{\textrm{L1}}(x)=
\begin{cases}
0.5x^2  & \text{if $ x < 1 $ } \\
|x|-0.5 & \text{otherwise}
\end{cases}
\end{equation}
and the temporal mask $g_m$ ($m \in \{l,a,v\}$) is used to exclude the loss from unmasked positions.

\subsubsection{High-Level Feature Attraction}
As a handcrafted pretext task, low-level feature reconstruction could fail to encourage the model to learn semantic information, as the model may find a shortcut to solve this task (i.e., only utilize local neighbor information to accomplish this task instead of inferring global semantics) \cite{jing2020self, bao2021beit, he2022masked}. To this end, we further utilize siamese representation learning to \textit{explicitly} attract high-level representations of complete and incomplete views in the latent space. As shown in Fig. \ref{fig_emt_dlfr}, except for $\tilde{X}_m$, we also pass $\mathring{X}_m$ through the model to get utterance-level inter-modal representation $\mathring{g}$ and intra-modal representations $\{\mathring{h}_m\}$ of complete view. 
 
The straightforward way to perform feature attraction is to maximize the cosine similarity of representations from two views. However, this method has the risk of collapsing as it may overwhelm the task loss and the model encodes both complete and incomplete inputs to constant vectors \cite{grill2020bootstrap, chen2021exploring}. 
Therefore, we employ a recently proposed self-supervised representation learning framework, SimSiam \cite{chen2021exploring}, to avoid possible collapsing solutions.
\begin{figure}[t]
	\centering
	\includegraphics[width=0.9\linewidth]{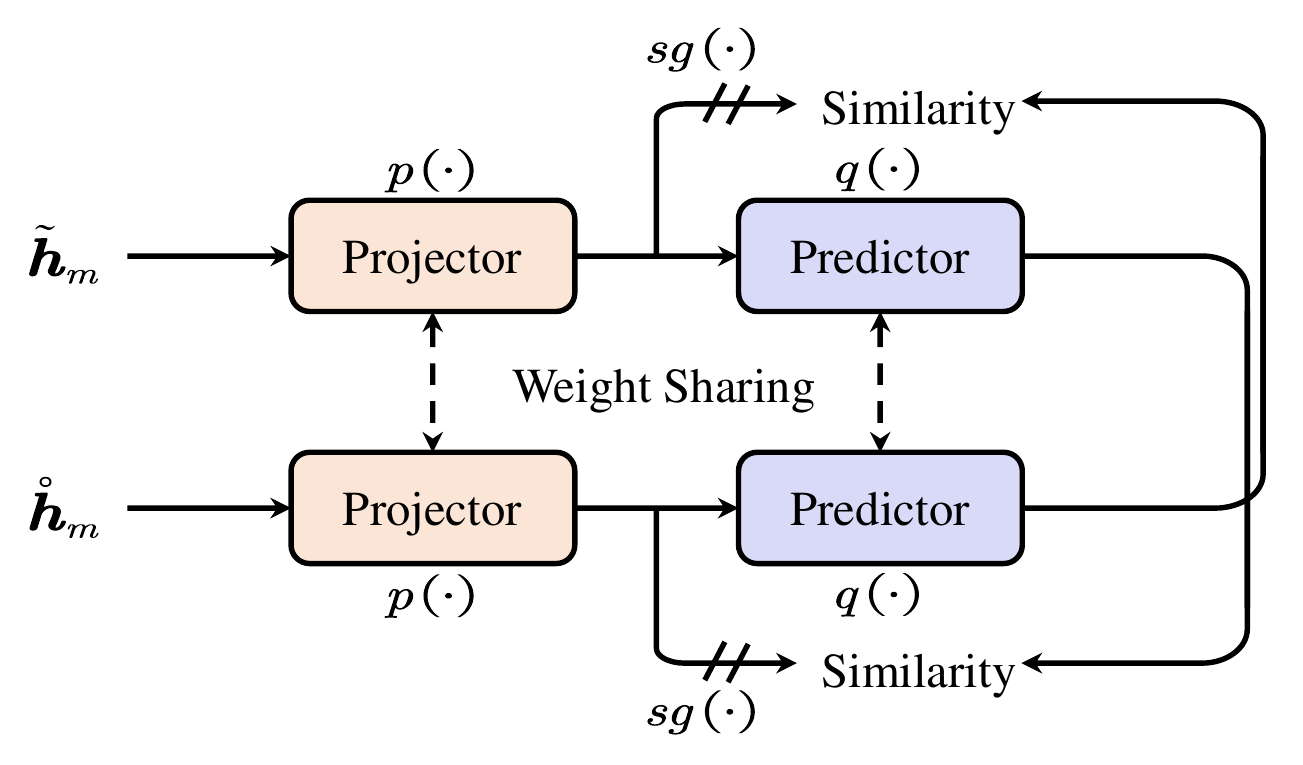}
	\caption{The illustration of symmetric SimSiam loss.}
	\label{fig_simsiam}
\end{figure}
As shown in Fig. \ref{fig_simsiam}, 
SimSiam consists of an MLP-based projector $p(\cdot)$ and an MLP-based predictor $q(\cdot)$. The input representations of two views are first processed by the projector. Then the predictor maps the intermediate representation of one view to that of the other view.
Besides, a stop-gradient operation $sg(\cdot)$ is used to avoid trivial solutions. Formally, we define the symmetric SimSiam loss of utterance-level intra- and inter-modal representations as follows:
\begin{equation}
\begin{split}
\mathcal{L}^m_{\textrm{sim}} &= \frac{1}{2}[\mathcal{D}(q(p(\tilde{h}_m)), sg(p(\mathring{h}_m))) + \mathcal{D}(q(p(\mathring{h}_m)), sg(p(\tilde{h}_m)))] \\
\mathcal{L}^g_{\textrm{sim}} &= \frac{1}{2}[\mathcal{D}(q(p(\tilde{g})), sg(p(\mathring{g}))) + \mathcal{D}(q(p(\mathring{g})), sg(p(\tilde{g})))]
\end{split}
\end{equation}
where,
\begin{equation}
 \mathcal{D}(x,y) = \frac{-x^Ty}{||x||_2||y||_2}
\end{equation}
is the negative cosine similarity function, and $m \in \{l, a, v\}$.
To ensure the quality of intra- and inter-modal representations from the complete view, we also send $\{\mathring{h}_m\}$ and $\mathring{g}$ to the prediction module and add the corresponding prediction loss in the attraction loss. Thus, the total feature attraction loss is as follows:
\begin{equation}
\mathcal{L}_{\textrm{attra}} = \sum_{m \in \{l, a, v\}}{\mathcal{L}^m_{\textrm{sim}}} + \mathcal{L}^g_{\textrm{sim}} +  |y - \mathring{y}'|
\label{eq_loss_hlfr}
\end{equation}

\subsection{Overall Loss Function}
\label{sec_method_loss}
In the incomplete modality setting, the task loss is the prediction loss of incomplete view, i.e., $\mathcal{L}_{\textrm{task}}=|y -\tilde{y}'|$. 
Thus, the overall loss function can be formulated as follows:
\begin{equation}
\begin{split}
\mathcal{L} &= \mathcal{L}_{\textrm{task}} + \lambda_1 \mathcal{L}_{\textrm{recon}} + \lambda_2  \mathcal{L}_{\textrm{attra}}
\end{split}
\label{eq_loss_overall}
\end{equation}
where $\lambda_1$ and $\lambda_2 \in \mathbb{R}$ are the weights that balance the contribution of dual-level feature restoration to $\mathcal{L}$.

In the complete modality setting, it is not necessary to perform dual-level feature restoration. 
Therefore, the overall loss function in this setting is $\mathcal{L} =\mathcal{L}_{\textrm{task}}=|y -\mathring{y}'|$.

\section{Experiments}
\label{sec_exp}
\subsection{Datasets}
In this paper, we conduct extensive experiments on three commonly used benchmark datasets in MSA. We give a brief introduction to each of them and summarize their basic statistics in Table \ref{tab_dataset_stat}.

\begin{table}[]
\caption{Statistics of three MSA benchmark datasets. The three numbers for each split denote the number of the samples with negative ($< 0$), neutral ($= 0$), and positive ($> 0$) sentiment, respectively.}
\label{tab_dataset_stat}
\resizebox{\linewidth}{!}{
\begin{tabular}{lrrrr}
\toprule
Dataset     & Train       & Val         & Test      & All \\
\midrule
CMU-MOSI    & 552/53/679       & 92/13/124        & 379/30/277     & 2199   \\
CMU-MOSEI   & 4738/3540/8048   & 506/433/932      & 1350/1025/2284 & 23453  \\
CH-SIMS     & 742/207/419      & 248/69/139       & 248/69/140     & 2281   \\
\bottomrule
\end{tabular}
}
\end{table}

\textbf{CMU-MOSI.}
The CMU-MOSI dataset \cite{zadeh2016multimodal} is a widely used MSA benchmark dataset in English. It is a collection of YouTube monologues in which the speakers share their opinions on a wide range of subjects (such as movies). CMU-MOSI consists of a total of 2,199 utterance-level video segments from 93 videos of 89 distinct speakers. Each video segment is manually annotated with a sentiment intensity score, which is defined from -3 (strongly negative) to 3 (strongly positive).

\textbf{CMU-MOSEI.}
The CMU-MOSEI dataset \cite{zadeh2018multimodal} is the next generation of CMU-MOSI.
Compared with CMU-MOSI, it has much larger training samples and more variations in speakers and video topics. 
Specifically, CMU-MOSEI contains 23,453 manually annotated utterance-level video segments from 1,000 distinct speakers and 250 different topics.

\textbf{CH-SIMS.}
The CH-SIMS dataset \cite{yu2020ch} is a Chinese MSA benchmark dataset. Compared with the above two datasets, it has both multimodal and unimodal annotations. Nevertheless, we only use the former in this paper. CH-SIMS consists of 2,281 utterance-level video segments from 60 videos whose types span from movies, TV series, and variety shows. Each video segment is manually annotated with a sentiment intensity score defined from -1 (strongly negative) to 1 (strongly positive).

\subsection{Evaluation Metrics}
\label{subsec_metric}
Since sentiment intensity prediction is primarily a regression task, the typically adopted evaluation metrics are mean absolute error (MAE) and Pearson correlation coefficient (Corr). 
Researchers also convert the continuous score into different discrete categories and report classification accuracy. Following previous works \cite{zadeh2018multimodal, tsai2019multimodal, hazarika2020misa, yuan2021transformer}, we report seven-class accuracy (Acc-7), five-class accuracy (Acc-5), binary accuracy (Acc-2), and F1-score on CMU-MOSI and CMU-MOSEI. 
Note that, there are two distinct methods for the binary formulation, i.e., negative/non-negative \cite{zadeh2018multimodal}, and negative/positive \cite{tsai2019multimodal}. Thus, we report the Acc-2 and F1-score of each method and use a segmentation marker -/- to differentiate them, where the left score is for negative/non-negative and the right score is for negative/positive.
On CH-SIMS, following \cite{yu2020ch}, we report five-class accuracy (Acc-5), three-class accuracy (Acc-3), and binary accuracy (Acc-2).
For all metrics but MAE, higher values indicate better performance.

To evaluate the model's overall performance under various missing rates in the incomplete modality setting, we follow \cite{yuan2021transformer} to compute the Area Under Indicators Line Chart (AUILC) for each of the above metrics. 
Suppose that the model performance on a metric under increasing missing rates $\{p_0, p_2, ..., p_t\}$ is $\{v_0, v_2, ..., v_t\}$, the AUILC of this metric is defined as follows:
\begin{equation}
\textrm{AUILC} = \sum_{i=0}^{t-1}{\frac{1}{2}(v_i+v_{i+1})(p_{i+1} - p_i)}
\end{equation}
To be consistent with \cite{yuan2021transformer}, we evaluate the model under the missing rates of $\{0.1, 0.2, ..., 1.0\}$ on CMU-MOSI and CMU-MOSEI, and under the missing rates of $\{0.1, 0.2, ..., 0.5\}$ on the CH-SIMS dataset.

\subsection{Feature Extraction}
\label{subsec_feature}
To make fair comparisons, we use the official unaligned features which are provided by the corresponding benchmark datasets and adopted by top-performing MSA methods.

\textbf{Text Modality.}
Transformer-based pre-trained language models have achieved state-of-the-art performances on a wide range of tasks in natural language processing. In agreement with recent works \cite{yu2021learning, han2021improving, yuan2021transformer}, we employ the pre-trained BERT model from the open-source Transformers library \cite{wolf2020transformers} to encode raw text. Specifically, we use \textit{bert-base-uncased} model for CMU-MOSI and CMU-MOSEI and \textit{bert-base-chinese} model for CH-SIMS.

\textbf{Audio Modality.}
For CMU-MOSI and CMU-MOSEI, COVAREP \cite{degottex2014covarep} acoustic analysis framework is utilized to extract the low-level acoustic features, which mainly consists of pitch, glottal source parameters, and 12 Mel-frequency cepstral coefficients (MFCCs). For CH-SIMS, an audio and music analysis Python package, Librosa \cite{mcfee2015librosa}, is used to extract logarithmic fundamental frequency, 12 Constant-Q chromatograms, and 20 MFCCs.

\textbf{Vision Modality.}
For CMU-MOSI and CMU-MOSEI, Facet\footnote{\url{https://imotions.com/platform/}} is employed to extract 35 facial action units, which record facial muscle movements related to emotions. For CH-SIMS, OpenFace 2.0 \cite{baltrusaitis2018openface} facial behavior analysis toolkit is used to extract 17 facial action units, 68 facial landmarks, and several features related to the head and eyes.

\subsection{Implementation Details}
We implement the proposed model using the PyTorch \cite{paszke2019pytorch} framework.
To train the model, we utilize an Adam \cite{kingma2014adam} optimizer and adopt an  early-stopping strategy with the patience of 8 epochs. 
For the hyper-parameters tuning, we perform a random search. 
The detailed configurations on CMU-MOSI, CMU-MOSEI, and CH-SIMS are summarized in Table \ref{tab_dataset_hyper}.
In the complete modality setting, aligned with \cite{yu2021learning, han2021improving}, we run the model five times and report average results for each evaluation metric. While in the incomplete modality setting, we follow \cite{yuan2021transformer} and run the model three times to ensure a fair comparison.

\begin{table}[]
\caption{Detailed configurations on three MSA benchmark datasets.}
\label{tab_dataset_hyper}
\resizebox{\linewidth}{!}{
\begin{tabular}{lrrr}
\toprule
Hyper-parameter                         & CMU-MOSI  & CMU-MOSEI   & CH-SIMS   \\
\midrule
Batch size                             & 32        & 16          & 32        \\
Learning rate                          & 1e-3      & 1e-4        & 1e-3      \\
Learning rate of BERT                  & 5e-5      & 2e-5        & 2e-5      \\
Optimizer                             & Adam      & Adam        & Adam      \\
Early stop ($\#$ epochs)              & 8         & 8           & 8         \\
Gradient accumulation ($\#$ batches)  & 4         & 4           & 4         \\
Hidden unit size $d$ in EMT            & 128       & 128         & 32        \\
Stacked layers $L$ in EMT              & 3         & 2           & 4         \\
$\#$ attention heads                   & 4         & 4           & 4         \\
Expansion factor of FFN                & 4         & 4           & 4         \\
Embedding dropout                      & 0.0       & 0.0         & 0.0       \\
Attention dropout                      & 0.3       & 0.0         & 0.0       \\
FFN dropout                            & 0.1       & 0.0         & 0.0       \\
Loss weight $\lambda_1$                & 1.0       & 1.0         & 0.5        \\
Loss weight $\lambda_2$                & 1.0       & 1.0         & 0.5       \\
\bottomrule
\end{tabular}
}
\end{table}

\begin{table*}[]
\caption{Performance comparison on CMU-MOSI and CMU-MOSEI in the complete modality setting. $^\dag$: results from \cite{han2021improving}. $^\ddag$: results from \cite{sun2022learning}. $^\diamondsuit$: results from \cite{yuan2021transformer}. All other results are reproduced using publicly available source codes and original hyper-parameters under the same setting. We run each model five times and report average results. For all metrics but MAE, higher values indicate better performance.}
\label{tab_mosi_mosei_complete}
\resizebox{\linewidth}{!}{
\begin{tabular}{lrrrrrrrrrrrrrrrrrc}
\toprule
 & \multicolumn{6}{c}{CMU-MOSI} &  \multicolumn{6}{c}{CMU-MOSEI}  \\
\cmidrule(lr){2-7} \cmidrule(lr){8-13}
Models    & MAE ($\downarrow$)   & Corr ($\uparrow$) & Acc-7 ($\uparrow$) &  Acc-5 ($\uparrow$) & Acc-2 ($\uparrow$) & F1 ($\uparrow$) & MAE ($\downarrow$)   & Corr ($\uparrow$) & Acc-7 ($\uparrow$) &  Acc-5 ($\uparrow$) & Acc-2 ($\uparrow$) & F1 ($\uparrow$) \\
\midrule
TFN$^\dag$     & 0.901 & 0.698 & 34.9  & -     &    -/80.8 &    -/80.7           & 0.593 & 0.700 & 50.2  & -     &    -/82.5 &    -/82.1          \\
LMF$^\dag$     & 0.917 & 0.695 & 33.2  & -     &    -/82.5 &    -/82.4           & 0.623 & 0.677 & 48.0  & -     &    -/82.0 &    -/82.1          \\
MulT$^\dag$    & 0.861 & 0.711 & -     & -     & 81.5/84.1 & 80.6/83.9           & 0.580 & 0.703 & -     & -     &    -/82.5 &    -/82.3           \\
MISA$^\dag$    & 0.804 & 0.764 & -     & -     & 80.8/82.1 & 80.8/82.0           & 0.568 & 0.724 & -     & -     & 82.6/84.2 & 82.7/84.0           \\
Self-MM$^\dag$ & 0.712 & 0.795 & 45.8  & -     & 82.5/84.8 & 82.7/84.9           & 0.529 & 0.767 & 53.5  & -     & 82.7/85.0 & 83.0/84.9          \\
MMIM$^\dag$    & 0.700 & 0.800 & 46.7  & -     & 84.1/86.1 & 84.0/86.0           & 0.526 & 0.772 & 54.2  & -     & 82.2/86.0 & 82.7/86.0          \\
AMML$^\ddag$           & 0.723 & 0.792 & 46.3  & -     &    -/84.9 &    -/84.8       & 0.614 & 0.776 & 52.4  & -     &    -/85.3 &    -/85.2 \\ 
TFR-Net$^\diamondsuit$ & 0.754 & 0.783 & -     & 54.7  &    -/84.1 &    -/-          & -     & -     & -     & -     &    -/-    &    -/-            \\
\midrule
MulT             & 0.846 & 0.725 & 40.4  & 46.7  & 81.7/83.4 & 81.9/83.5         & 0.564 & 0.731 & 52.6  & 54.1  & 80.5/83.5 & 80.9/83.6            \\
Self-MM          & 0.717 & 0.793 & 46.4  & 52.8  & 82.9/84.6 & 82.8/84.6         & 0.533 & 0.766 & 53.6  & 55.4  & 82.4/85.0 & 82.8/85.0          \\
MMIM             & 0.712 & 0.790 & 46.9  & 53.0  & \textbf{83.3}/\textbf{85.3} & \textbf{83.4}/\textbf{85.4}  & 0.536 & 0.764 & 53.2  & 55.0  & 82.5/85.0 & 82.4/85.1  \\
TFR-Net          & 0.721 & 0.789 & 46.1  & 53.2  & 82.7/84.0 & 82.7/84.0         & 0.551 & 0.756 & 52.3  & 54.3  & 81.8/83.5 & 81.6/83.8            \\
EMT              & \textbf{0.705} & \textbf{0.798} & \textbf{47.4}  & \textbf{54.1}  & \textbf{83.3}/85.0 & 83.2/85.0              & \textbf{0.527} & \textbf{0.774} & \textbf{54.5}  & \textbf{56.3}  & \textbf{83.4}/\textbf{86.0} & \textbf{83.7}/\textbf{86.0}                        \\
\bottomrule
\end{tabular}
}
\end{table*}

\begin{table}[]
\caption{Performance comparison on CH-SIMS in the complete modality setting. $^\dag$: results from \cite{mao2022m}. All other results are reproduced using publicly available source codes and original hyper-parameters under the same setting. We run each model five times and report average results. For all metrics but MAE, higher values indicate better performance.}
\label{tab_sims_complete}
\resizebox{\linewidth}{!}{
\begin{tabular}{lrrrrrr}
\toprule
Models    & MAE ($\downarrow$)   & Corr ($\uparrow$) & Acc-5 ($\uparrow$) & Acc-3 ($\uparrow$) & Acc-2 ($\uparrow$) & F1 ($\uparrow$) \\
\midrule
TFN$^\dag$      & 0.437 & 0.582 & -     & -     & 77.1  & 76.9 \\
LMF$^\dag$      & 0.438 & 0.578 & -     & -     & 77.4  & 77.4 \\
MulT$^\dag$     & 0.453 & 0.564 & -     & -     & 78.6  & 79.7 \\
MISA$^\dag$     & 0.447 & 0.563 & -     & -     & 76.5  & 76.6 \\
Self-MM$^\dag$  & 0.425 & 0.595 & -     & -     & 80.0  & 80.4 \\
\midrule
MulT             & 0.442 & 0.581 & 40.0  & 65.7  & 78.2  & 78.5\\
Self-MM          & 0.411 & 0.601 & 43.1  & 66.1  & 78.6  & 78.6\\
MMIM             & 0.422 & 0.597 & 42.0  & 65.5  & 78.3  & 78.2\\
TFR-Net          & 0.437 & 0.583 & 41.2  & 64.2  & 78.0  & 78.1\\
EMT       & \textbf{0.396} & \textbf{0.623} & \textbf{43.5}  & \textbf{67.4}  & \textbf{80.1} & \textbf{80.1} \\ 
\bottomrule
\end{tabular}
}
\end{table}

\subsection{Baselines}
To comprehensively evaluate the performance of our proposed method in incomplete and complete settings, we consider both general MSA approaches and those which are specially designed to deal with random modality feature missing as our baselines. 

\textbf{TFN}. Tensor Fusion Network (TFN) \cite{zadeh2017tensor} introduces a three-fold Cartesian product-based tensor fusion layer to explicitly model intra-modality and inter-modality dynamics in an end-to-end manner.

\textbf{LMF}. Low-rank Multimodal Fusion (LMF) \cite{liu2018efficient} leverages modality-specific low-rank factors to compute tensor-based multimodal representations, which makes tensor fusion more efficient.

\textbf{MulT}. Multimodal Transformer (MulT) \cite{tsai2019multimodal} utilizes directional pairwise cross-modal attention to capture inter-modal correlations in unaligned multimodal sequences.

\textbf{MISA}. This method \cite{hazarika2020misa} factorizes modalities into Modality-Invariant and -Specific Representations (MISA) using a combination of specially designed losses and then performs multimodal fusion on these representations.

\textbf{Self-MM}. Self-Supervised Multi-task Multimodal (Self-MM) sentiment analysis network \cite{yu2021learning} designs a unimodal label
generation module based on self-supervised learning to explore unimodal supervision.

\textbf{MMIM}. MultiModal InfoMax (MMIM) \cite{han2021improving} proposes a hierarchical mutual information maximization framework to guide the model to learn shared representations from all modalities.

\textbf{AMML}. Adaptive Multimodal Meta-Learning (AMML) \cite{sun2022learning} introduces a meta-learning-based method to learn better unimodal representations and then adapt them for subsequent multimodal fusion.

\textbf{TFR-Net}. Transformer-based Feature Reconstruction Network (TFR-Net) \cite{yuan2021transformer} employs intra- and inter-modal attention and a feature reconstruction module to deal with random modality feature missing in unaligned multimodal sequences.

\section{Results}

\subsection{Comparison to State-of-the-art}

\subsubsection{Complete Modality Setting} 
First, we compare the proposed EMT with top-performing MSA methods in the complete modality setting. 
Table \ref{tab_mosi_mosei_complete} shows the results on CMU-MOSI and CMU-MOSEI. 
Since the baseline papers do not have a common evaluation setting and only report results on partial metrics, we present both the results from original papers and those reproduced by ourselves under the same setting (they are generally comparable). For fair and full comparisons, we mainly compare ours with the reproduced one.
From Table \ref{tab_mosi_mosei_complete}, we first observe that our EMT shows superior performance over previous state-of-the-art Transformer-based methods (i.e., TFR-Net, and MulT). 
Specifically, it surpasses the best-performing TFR-Net by 0.016 MAE on CMU-MOSI and 2.2\% Acc-7 on CMU-MOSEI, and outperforms MulT by 0.141 MAE on CMU-MOSI and 0.043 Corr on CMU-MOSEI.
We can attribute these encouraging results to the effective yet efficient global-local cross-modal interaction modeling in EMT, because both two baseline methods focus on capturing pairwise local-local cross-modal interactions, which not only suffer from a large amount of redundancy but also increase the risk of overfitting.
In addition to Transformer-based methods, EMT also surpasses several recent state-of-the-art non-Transformer-based methods (e.g., AMML, MMIM, and Self-MM), setting new  records on most metrics.
Finally, we present the results on the Chinese MSA dataset CH-SIMS in Table \ref{tab_sims_complete}.
It can be seen that EMT achieves the best performance on all metrics. For instance, it outperforms the second performer Self-MM by 0.015 MAE, 0.022 Corr, and 1.5\% Acc-2. 
In summary, the above results demonstrate the effectiveness of EMT in the complete modality setting.

\subsubsection{Incomplete Modality Setting} 

\begin{figure*}[htbp]
	\centering
	\includegraphics[width=\linewidth]{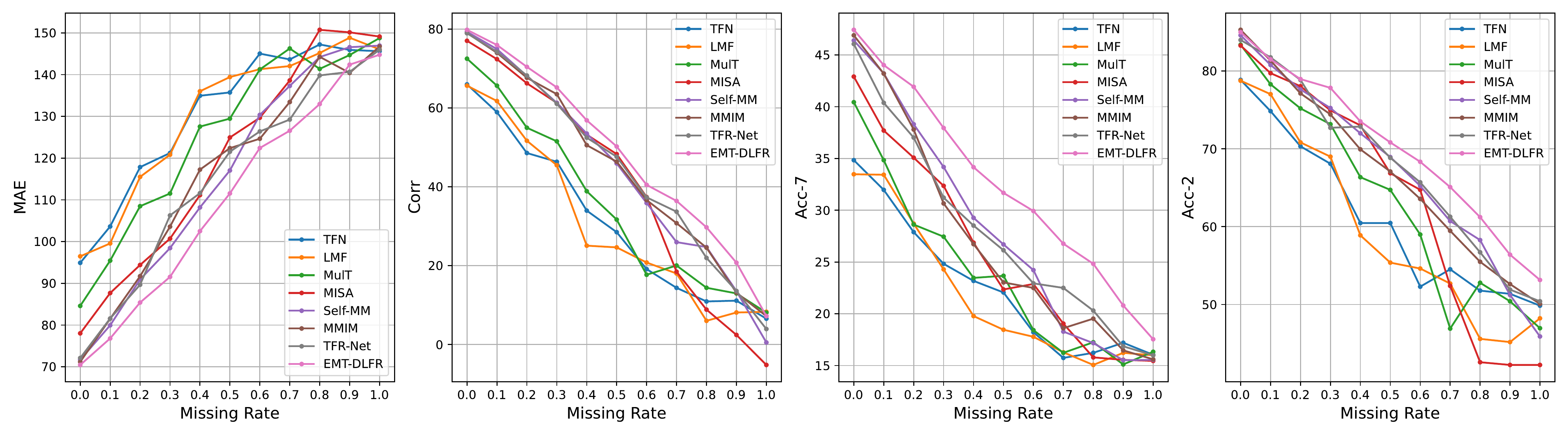}
	\caption{Performance comparison under various missing rates on CMU-MOSI.}
	\label{fig_mosi_incomplete}
\end{figure*}

\begin{figure*}[t]
	\centering
	\includegraphics[width=\linewidth]{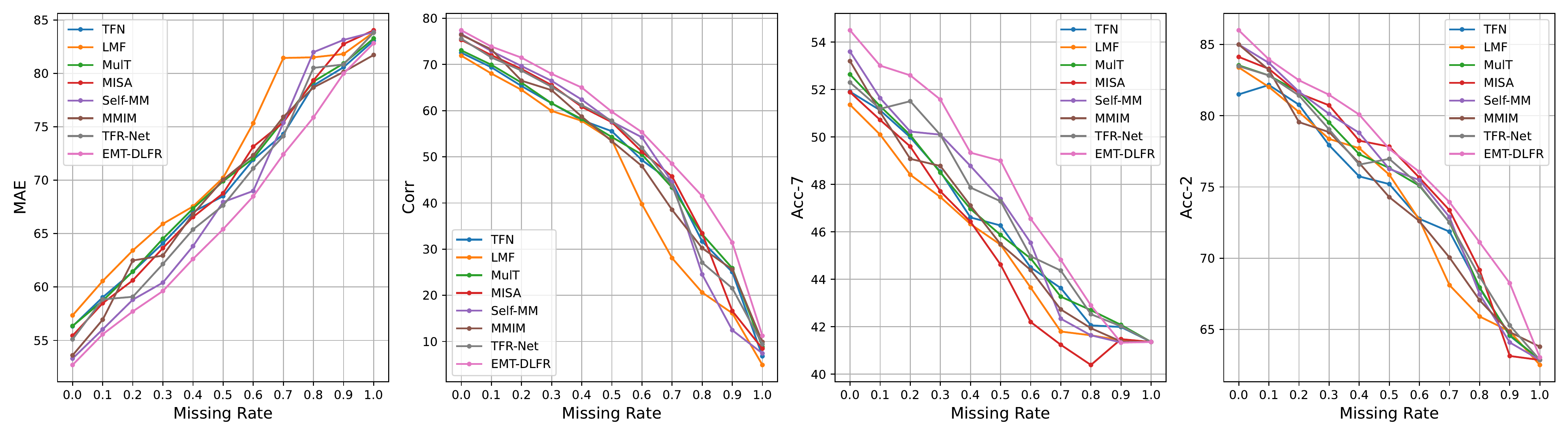}
	\caption{Performance comparison under various missing rates on CMU-MOSEI.}
	\label{fig_mosei_incomplete}
\end{figure*}

\begin{figure*}[t]
	\centering
	\includegraphics[width=\linewidth]{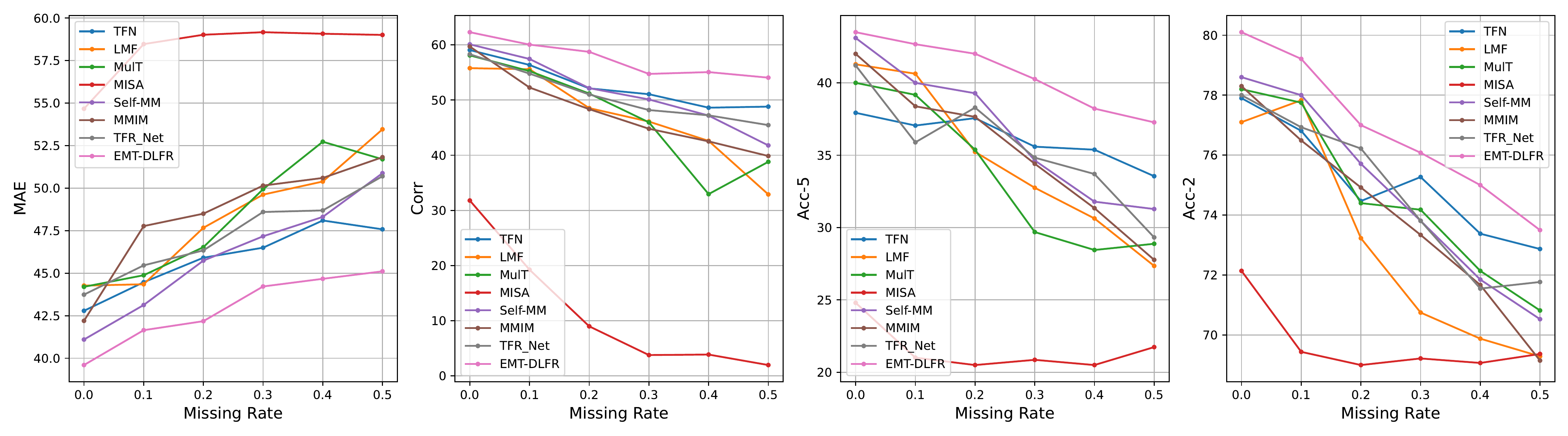}
	\caption{Performance comparison under various missing rates on CH-SIMS.}
	\label{fig_sims_incomplete}
\end{figure*}

\begin{table*}[]
\caption{Overall performance comparison on CMU-MOSI and CMU-MOSEI in the incomplete modality setting. The reported result is the AUILC of each evaluation metric, which is calculated under the missing rates of $\{0.1, 0.2, ..., 1.0\}$. $^\diamondsuit$: results from \cite{yuan2021transformer}. All other results are reproduced using publicly available source codes and original hyper-parameters under the same setting. We run each model three times and report average results. For all metrics but MAE, higher values indicate better performance.}
\label{tab_mosi_mosei_incomplete}
\resizebox{\linewidth}{!}{
\begin{tabular}{lrrrrrrrrrrrrrrrrrc}
\toprule
 & \multicolumn{6}{c}{CMU-MOSI} &  \multicolumn{6}{c}{CMU-MOSEI}  \\
\cmidrule(lr){2-7} \cmidrule(lr){8-13}
Models    & MAE ($\downarrow$)   & Corr ($\uparrow$) & Acc-7 ($\uparrow$) &  Acc-5 ($\uparrow$) & Acc-2 ($\uparrow$) & F1 ($\uparrow$) & MAE ($\downarrow$)   & Corr ($\uparrow$) & Acc-7 ($\uparrow$) &  Acc-5 ($\uparrow$) & Acc-2 ($\uparrow$) & F1 ($\uparrow$)       \\
\midrule
TFN$^\diamondsuit$       & 1.327 & 0.300 & -     & 23.3  & -/60.4    & -/-       & - & - & -     & -  & -/-    & -/-          \\
MulT$^\diamondsuit$      & 1.288 & 0.334 & -     & 24.4  & -/61.8    & -/-       & - & - & -     & -  & -/-    & -/-            \\
MISA$^\diamondsuit$      & 1.209 & 0.403 & -     & 27.1  & -/63.2    & -/-       & - & - & -     & -  & -/-    & -/-        \\
TFR-Net$^\diamondsuit$   & 1.155 & 0.467 & -     & 30.4  & -/69.0    & -/-       & - & - & -     & -  & -/-    & -/-        \\
\midrule
TFN              & 1.316 & 0.308 & 22.3  & 23.7  & 61.0/60.9 & 59.7/59.7      & 0.695 & 0.500 & 46.1  & 46.6  & 75.2/74.1 & 73.4/71.5      \\
LMF              & 1.310 & 0.299 & 21.5  & 22.7  & 59.7/59.3 & 56.4/56.1      & 0.718 & 0.447 & 45.3  & 45.7  & 72.2/73.9 & 69.5/69.4      \\
MulT             & 1.263 & 0.348 & 23.1  & 24.6  & 63.1/63.2 & 60.7/61.0      & 0.700 & 0.504 & 46.3  & 46.8  & 74.4/75.1 & 72.9/72.6      \\
MISA             & 1.202 & 0.405 & 25.7  & 27.4  & 63.9/63.7 & 59.0/58.8      & 0.698 & 0.514 & 45.1  & 45.7  & 75.2/75.7 & 74.4/74.0       \\
Self-MM          & 1.162 & 0.444 & 27.8  & 30.3  & 66.9/67.5 & 65.4/66.2      & 0.685 & 0.507 & 46.7  & 47.3  & 75.1/75.4 & 73.7/72.9       \\ 
MMIM             & 1.168 & 0.450 & 27.0  & 29.4  & 66.8/66.9 & 64.6/65.8      & 0.694 & 0.502 & 45.9  & 46.4  & 74.9/72.4 & 72.4/69.3       \\ 
TFR-Net          & 1.156 & 0.452 & 27.7  & 30.5  & 67.6/67.8 & 65.7/66.1      & 0.689 & 0.511 & 46.9  & 47.3  & 74.7/74.2 & 73.5/73.4       \\
EMT-DLFR  & \textbf{1.106} & \textbf{0.486} & \textbf{32.5}  & \textbf{35.6}  & \textbf{69.6}/\textbf{70.3} & \textbf{69.6}/\textbf{70.3}          & \textbf{0.665} & \textbf{0.546} & \textbf{47.9}  & \textbf{48.8}  & \textbf{76.4}/\textbf{76.9} & \textbf{75.2}/\textbf{75.9}         \\
\bottomrule
\end{tabular}
}
\end{table*}

\begin{table}[]
\caption{Overall performance comparison on CH-SIMS in the incomplete modality setting. The reported results are the AUILC of each evaluation metric, which are calculated under the missing rates of $\{0.1, 0.2, ..., 0.5\}$. $^\diamondsuit$: results from \cite{yuan2021transformer}. All other results are reproduced using publicly available source codes and original hyper-parameters under the same setting. We run each model three times and report average results. For all metrics but MAE, higher values indicate better performance.}
\label{tab_sims_incomplete}
\resizebox{\linewidth}{!}{
\begin{tabular}{lrrrrrr}
\toprule
Models    & MAE ($\downarrow$)   & Corr ($\uparrow$) & Acc-5 ($\uparrow$) & Acc-3 ($\uparrow$) & Acc-2 ($\uparrow$) & F1 ($\uparrow$)\\
\midrule
TFN$^\diamondsuit$       & 0.233 & 0.259 & 18.1  & -     & 37.3  & -   \\
MulT$^\diamondsuit$      & 0.244 & 0.227 & 17.3  & -     & 37.0  & -   \\
MISA$^\diamondsuit$      & 0.294 & 0.038 & 10.6  & -     & 34.7  & -   \\
TFR-Net$^\diamondsuit$   & 0.237 & 0.253 & 18.0  & -     & 37.7  & -   \\
\midrule
TFN              & 0.230 & 0.262 & 18.1  & 30.5  & 37.5  & 37.5  \\
LMF              & 0.241 & 0.237 & 17.4  & 30.1  & 36.5  & 36.1  \\
MulT             & 0.242 & 0.233 & 16.7  & 30.0  & 37.3  & 36.8  \\
MISA             & 0.293 & 0.053 & 10.6  & 26.4  & 34.8  & 28.9  \\
Self-MM          & 0.231 & 0.258 & 18.3  & 30.5  & 37.4  & 37.5   \\
MMIM             & 0.244 & 0.238 & 17.7  & 29.8  & 37.0  & 36.2   \\
TFR-Net          & 0.236 & 0.253 & 17.8  & 30.0  & 37.3  & 37.2   \\
EMT-DLFR  & \textbf{0.215} & \textbf{0.287} & \textbf{20.4}  & \textbf{31.9}  & \textbf{38.4} & \textbf{38.5} \\
\bottomrule
\end{tabular}
}
\end{table}

Next, we concentrate on evaluating the robustness of our EMT-DLFR under various missing rates. Note that, to generate incomplete views, we apply the same missing rate to three modalities during both the training and test stages.
Besides, we independently generate the random temporal mask for each modality.
The performance comparison on CMU-MOSI, CMU-MOSEI, and CH-SIMS are shown in Fig. \ref{fig_mosi_incomplete}, \ref{fig_mosei_incomplete}, and \ref{fig_sims_incomplete}, respectively. 
For simplicity, we only show four representative metrics (i.e., MAE, Corr, Acc-7/Acc-5, and Acc-2) for each dataset. Note that the Acc-2 on CMU-MOSI and CMU-MOSEI means the binary accuracy calculated on negative/positive samples.
From the three figures, we have the following observations:
(1) In general, with the increase in missing rates, the performances of all methods decrease quasi-linearly. It indicates that random modality feature missing in unaligned multimodal sequences will moderately degrade the model's performance and this issue needs to be carefully addressed in the real-world deployment of MSA models.
(2) EMT-DLFR outperforms all other compared methods in most cases on three datasets, especially when the missing rate is relatively high, thus clearly illustrating the robustness of our proposed method in the incomplete modality setting.

To quantitatively evaluate the overall performance of different methods, we follow \cite{yuan2021transformer} to calculate the AUILC of each evaluation metric, as stated in Section \ref{subsec_metric}. 
The results on CMU-MOSI and CMU-MOSEI are shown in Table \ref{tab_mosi_mosei_incomplete}. 
Similar to the complete case, we report both original and reproduced results, and we mainly compare ours with the latter. 
On CMU-MOSI, we can find that the state-of-the-art TFR-Net which is specialized to deal with random modality feature missing outperforms other general MSA methods in most cases. This indicates that low-level feature reconstruction adopted by TFR-Net can indeed force the model to learn certain semantic information from incomplete multimodal sequences. Nevertheless, our EMT-DLFR which utilizes both low-level feature reconstruction and high-level feature attraction improves both TFR-Net and other methods by a large margin on almost all metrics. To be specific, it outperforms the second performer by 0.050 MAE, 0.034 Corr, 4.7\%  Acc-7, 5.1\% Acc-5, and 3.9\%/4.1\% F1, which amply demonstrates the superiority of the proposed method.
On CMU-MOSEI, we observe that the difference between different methods is smaller than that on CMU-MOSI (also shown in Fig. \ref{fig_mosei_incomplete}), which could be partly ascribed to much larger training samples in this dataset. Our EMT-DLFR still achieves 0.020 and 0.032 higher performance on MAE and Corr than the second performer.
We further present the results on CH-SIMS in Table \ref{tab_sims_incomplete}. The proposed method also achieves superior performance, outperforming the previous state-of-art by 0.015 MAE and 0.025 Corr.
To summarize, the above results verify the robustness of our proposed method in the incomplete modality setting.

\subsection{Ablation Study} 
\label{sec_ablation}
To further investigate the influence of each component in our method, we conduct comprehensive ablation experiments on CMU-MOSI in the incomplete modality setting.

\subsubsection{Multimodal Fusion Strategy} 

\begin{figure}[t]
	\centering
	\includegraphics[width=\linewidth]{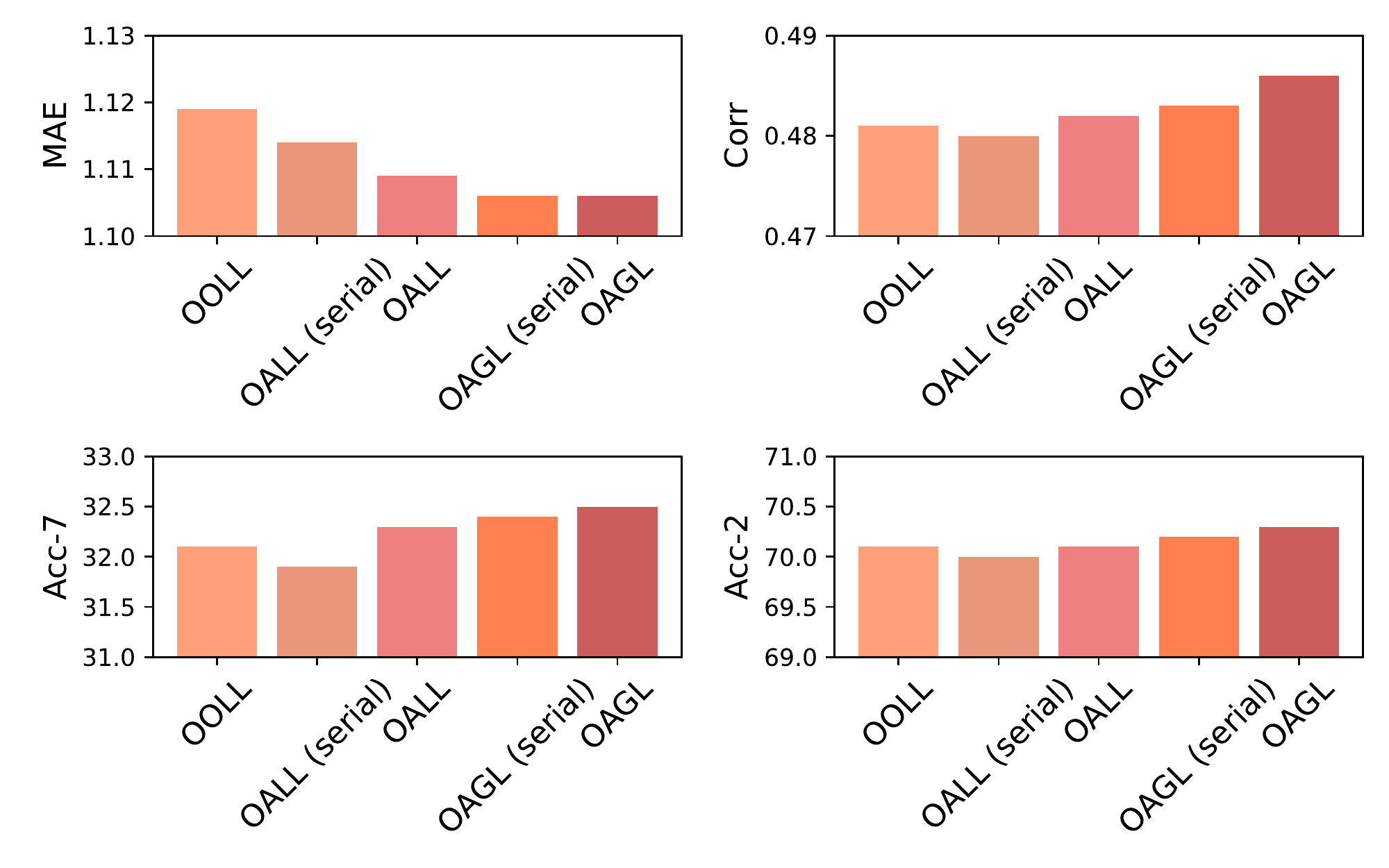}
	\caption{Ablation study of multimodal fusion strategy on CMU-MOSI in the incomplete modality setting.}
	\label{fig_mosi_ablation_fusion}
\end{figure}

We first validate the effectiveness and efficiency of the proposed multimodal fusion strategy. We compare the default OAGL fusion strategy adopted by EMT with two baseline fusion strategies (i.e., OOLL and OALL) inspired by previous methods as stated in Section \ref{sec_method_emt}. 
We show the performance of three fusion strategies in Fig. \ref{fig_mosi_ablation_fusion}.
We can find that OAGL outperforms OOLL and OALL, which verifies the feasibility and effectiveness of utilizing utterance-level representations from each modality as the global multimodal context to interact with local unimodal features. 
We think that the superiority of OAGL mainly comes from the simplification of cross-modal interaction modeling. In comparison to the local-local version in OOLL and OALL, it not only minimizes redundant information but also encourages the model to concentrate exclusively on prominent cross-modal correlation patterns, thus making it less likely to overfit the spurious ones.
Additionally, inspired by \cite{lv2021progressive}, for OALL and OAGL, we develop their serial variants by replacing $H_{m}$ in the last row of Equation \ref{eq_mpu} with $H_{n \rightarrow m}$ (i.e., the promoted $H_{m}$) in the second row. From Fig. \ref{fig_mosi_ablation_fusion}, we observe that the serial variant performs a bit worse than the default one for both OALL and OAGL, which justifies the parallel design choice of two fusion strategies. We believe this might be due to the loss of original unimodal information brought by the too-fast feature update in the serial implementation.

To further illustrate the efficiency of OAGL over OOLL and OALL, we give a practical complexity comparison of them. Since there are three input modalities, these three strategies have the same space complexity (as they all need 3 MPUs in each fusion layer). Thus, we mainly evaluate the time complexity in this paper. Nevertheless, it is worth noting that OAGL has linear scalability of MPUs when more modalities are involved in the fusion process.
In Table \ref{tab_fusion_efficiency}, we compare three fusion strategies in terms of the amount of computation measured in multiply–accumulate operations (MACs) \footnote{\url{https://github.com/Lyken17/pytorch-OpCounter}}, the number of parameters, one epoch training time, and GPU memory usage during model training. 
Note that the MACs are only calculated on the fusion module, excluding the other parts (e.g., unimodal feature encoders) of the model. For the number of parameters, we do not count the networks for feature reconstruction and attraction because they will not be used during inference. We run each experiment five times using a batch size of 32 on a single Tesla V100 GPU (32GB) and report the mean value for each metric. In addition to OOLL and OALL, we also provide metrics of two state-of-the-art multimodal Transformers as references, including MulT and TFR-Net.

From Table \ref{tab_fusion_efficiency}, we can observe that OAGL has the least MACs among the three fusion strategies (5$\times$ less than OALL and 2$\times$ less than OOLL). This is expected since OAGL enjoys linear computational complexity over the involved modalities in practice while both OOLL and OALL have a quadratic scaling cost.
Compared with OOLL, OALL which utilizes the common message (i.e., local multimodal context) to explore multi-way cross-modal interactions achieves better performance (as shown in Fig. \ref{fig_mosi_ablation_fusion}) but at the expense of a much larger amount of computation (especially the GPU memory usage). In contrast, thanks to the introduction of the global multimodal context, our OAGL not only has the best performance but also has the least computation. 
We also notice that the improvement in training speed is less significant than that of MACs. This is owing to the dominant role of the text modality feature encoder (i.e., BERT) in the overall computational overhead. 
Finally, the state-of-the-art TFR-Net and MulT, which utilize similar strategies to OOLL for multimodal fusion, also have much larger computational costs and more parameters than our OAGL. 
To summarize, the above quantitative results demonstrate the efficiency of OAGL over traditional local-local methods for cross-modal interaction modeling.

\begin{table}[]
\caption{The complexity comparison of different multimodal fusion strategies on CMU-MOSI in the incomplete modality setting.}
\label{tab_fusion_efficiency}
\resizebox{\linewidth}{!}{
\begin{tabular}{lcccc}
\toprule
\tabincell{l}{Fusion\\Strategy}    & \tabincell{c}{MACs\\(G)} & \tabincell{c}{\#Params\\(M)}  & \tabincell{c}{Training Time\\(s)} & \tabincell{c}{GPU Memory\\(GB)} \\
\midrule
MulT        & -    & 111.0 & 17.5  & 17.8  \\
TFR-Net     & -    & 124.3 & 24.8  & 16.9   \\
\midrule
OOLL        & 3.1  & \textbf{110.5} & 17.1  & 17.8  \\
OALL        & 8.3  & \textbf{110.5} & 21.2  & 31.5  \\
OAGL        & \textbf{1.5}  & \textbf{110.5} & \textbf{15.4}  & \textbf{10.8} \\
\bottomrule
\end{tabular}
}
\end{table}

\subsubsection{Hierarchical Parameter Sharing for EMT}
To verify the role of hierarchical parameter sharing for EMT, we evaluate different combinations of three-level (i.e., MPU-level, modality-level, and layer-level) parameter sharing strategies in Table \ref{tab_ablation_param_sharing}.
Surprisingly, we find that all methods have comparable performance and the full (i.e., not-shared) model is even slightly inferior to several parameter-sharing variants. Moreover, the all-shared model achieves decent results only with negligible performance degradation on MAE when compared to the full model. 
These results indicate that EMT is more difficult to train when there are too many parameters, and parameter sharing can act as a regularization constraint to alleviate this problem to some extent.
Another benefit of parameter sharing is the big improvement in parameter efficiency.
Note that, compared with the not-shared case, sharing all parameters in EMT leads to a reduction of 6M parameters. Since the text BERT encoder has 109.5M parameters, there are only less than 1M (actually, 0.5M) parameters in EMT for the all-shared model.
\begin{table}[]
\caption{Ablation study of hierarchical parameter sharing in EMT on CMU-MOSI in the incomplete modality setting.}
\label{tab_ablation_param_sharing}
\resizebox{\linewidth}{!}{
\begin{tabular}{cccccccc}
\toprule
MPU  & Modality  & Layer & \#Params (M)   & MAE ($\downarrow$)  & Corr ($\uparrow$)  \\
\midrule
$\times$      & $\times$   &  $\times$  &  116.4             & 1.105 & 0.486  \\
\checkmark    & $\times$   &  $\times$  &  113.4             & 1.103 & 0.484  \\
$\times$      & \checkmark &  $\times$  &  112.4             & 1.105 & \textbf{0.489}  \\
$\times$      & $\times$   & \checkmark &  112.1             & 1.104 & 0.487  \\
\checkmark    & \checkmark &  $\times$  &  111.4             & 1.103 & 0.487  \\
\checkmark    & $\times$   & \checkmark &  111.1             & 1.106 & 0.485  \\
$\times$      & \checkmark & \checkmark &  110.8             & \textbf{1.101} & 0.488  \\
\checkmark    & \checkmark & \checkmark &  \textbf{110.5}    & 1.106 & 0.486  \\
\bottomrule
\end{tabular}
}
\end{table}

\subsubsection{Pooling Layer in EMT}
\begin{figure}[t]
	\centering
	\includegraphics[width=\linewidth]{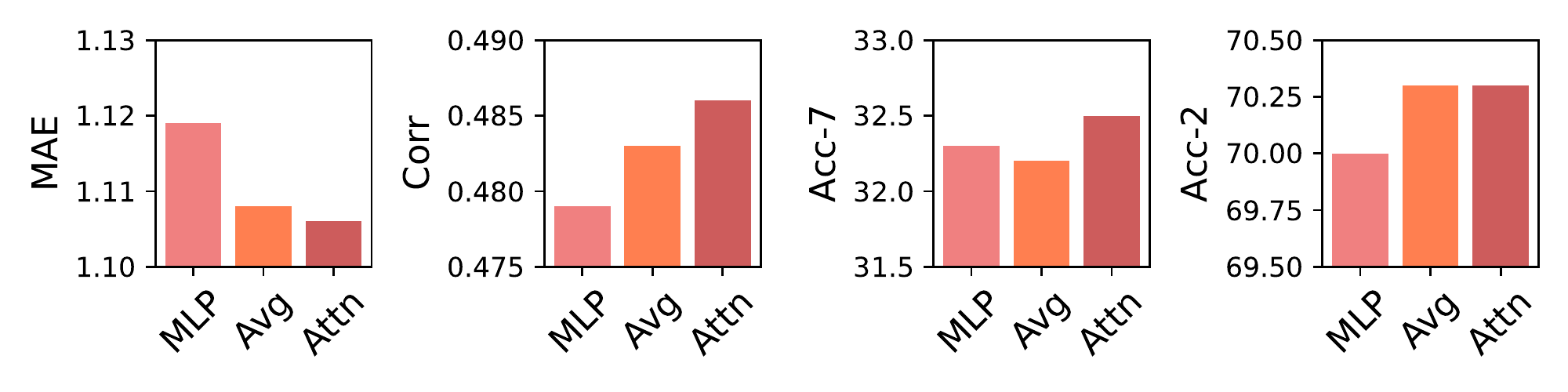}
	\caption{Ablation study of pooling layer in EMT on CMU-MOSI in the incomplete modality setting.}
	\label{fig_mosi_ablation_pooling}
\end{figure}

In this part, we ablate the design choice of the pooling layer in EMT.
The pooling layer in EMT is used to aggregate promoted information from different modalities.
We compare the default attention pooling with two variants, i.e., average pooling and MLP-based pooling. The results are shown in Fig. \ref{fig_mosi_ablation_pooling}. As a non-parametric method, average pooling is slightly inferior to attention pooling since it can not make aware of the contributions of different modalities. MLP-based pooling directly maps multiple promoted features into a single one using a fully connected layer. Thus, compared with attention-pooling, it has much more parameters. However, MLP-based pooling achieves the worst performance among the three methods. We guess this could be caused by the overfitting problem considering that MSA datasets are relatively small.

\subsubsection{Dual-level Feature Restoration}
We then investigate the contributions of low-level feature reconstruction and high-level feature attraction to the improvement of model robustness. We compare our full method EMT-DLFR with the following three variants: 
\begin{itemize}
\item EMT-LLFR: only use low-level feature reconstruction by removing $\mathcal{L}_{\textrm{attra}}$ in Equation \ref{eq_loss_overall}.
\item EMT-HLFR: only use high-level feature attraction by removing $\mathcal{L}_{\textrm{recon}}$ in Equation \ref{eq_loss_overall}.
\item EMT: use neither low-level feature reconstruction nor high-level feature attraction by removing both $\mathcal{L}_{\textrm{recon}}$ and $\mathcal{L}_{\textrm{attra}}$ in Equation \ref{eq_loss_overall}. 
\end{itemize}

\begin{figure}[t]
	\centering
	\includegraphics[width=\linewidth]{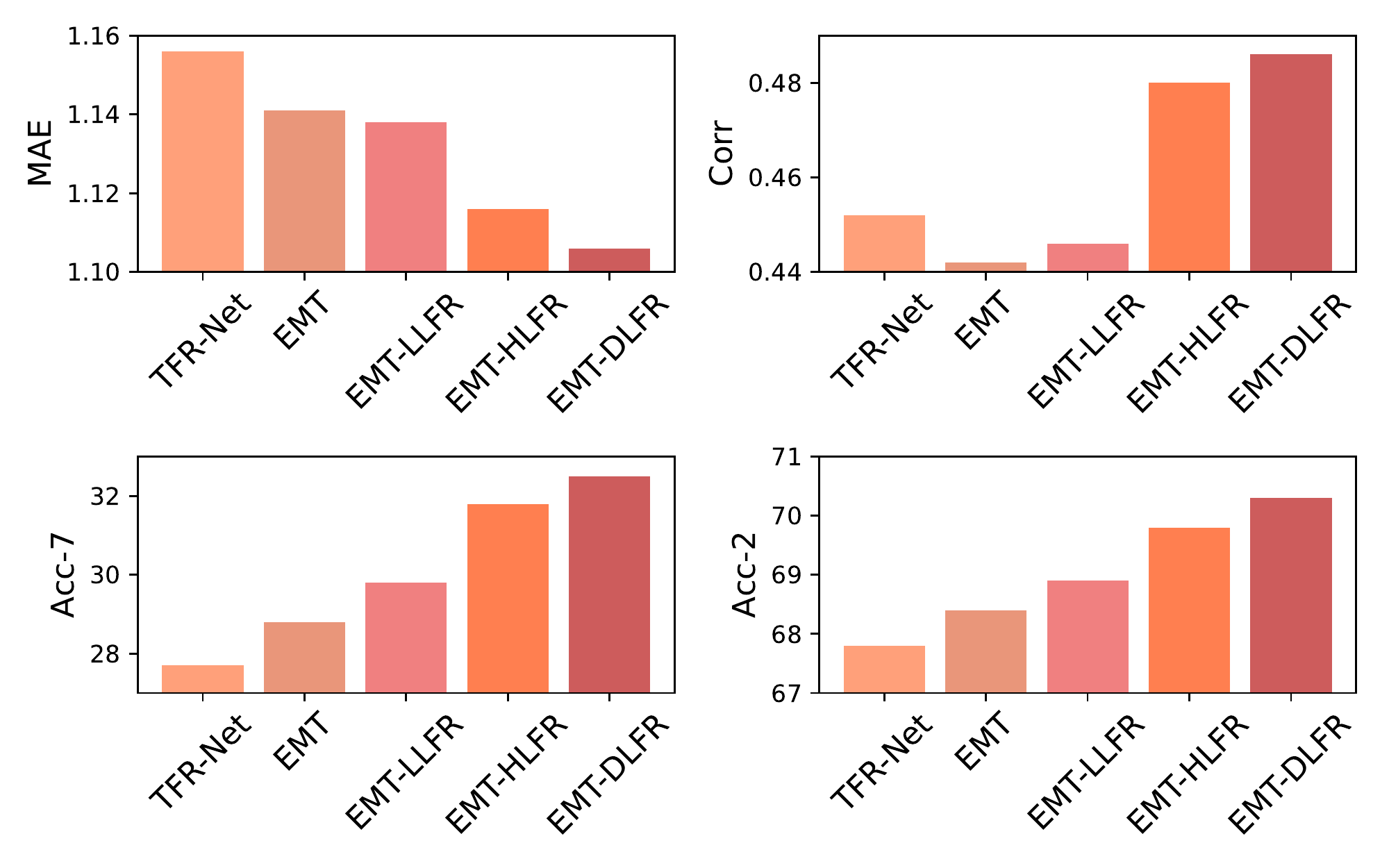}
	\caption{Ablation study of loss function on CMU-MOSI in the incomplete modality setting.}
	\label{fig_mosi_ablation_loss}
\end{figure}

Fig. \ref{fig_mosi_ablation_loss} shows the results of different methods. We can observe that EMT-LLFR and EMT-HLFR outperform vanilla EMT, which indicates both low-level feature reconstruction and high-level feature attraction can improve model robustness to random modality feature missing. Besides, high-level feature attraction is more crucial than low-level feature reconstruction considering that EMT-LLFR significantly underperforms EMT-HLFR by 0.022 MAE, 0.034 Corr, 2.0\% Acc-7, and 0.9\% Acc-2. This is conceptually intuitive since implicit low-level feature reconstruction might not be sufficient to force the model to infer high-level semantics,  as it could find a shortcut (i.e., only using local neighbor information) to accomplish the reconstruction task. 
In contrast, directly performing explicit attraction of high-level features from the complete and incomplete view in the latent space is more beneficial and effective. Moreover, our full method EMT-DLFR achieves the best performance among all methods, which suggests that implicit low-level feature reconstruction and explicit high-level feature attraction are complementary to each other. Therefore, the combination of these two feature restoration mechanisms can be a unified framework for robust multimodal sentiment analysis.
For reference, we also present the result of state-of-the-art TFR-Net in Fig. \ref{fig_mosi_ablation_loss}. As expected, both EMT-LLFR and EMT-HLFR have better performance than TFR-Net. The encouraging thing is that even vanilla EMT slightly outperforms TFR-Net on several metrics (e.g., MAE, Acc-7, and Acc-2), which once again demonstrates the superiority of EMT over traditional local-local cross-modal fusion methods.

\subsubsection{Sensitivity of Loss Weights}
\begin{figure}[t]
	\centering
	\includegraphics[width=0.9\linewidth]{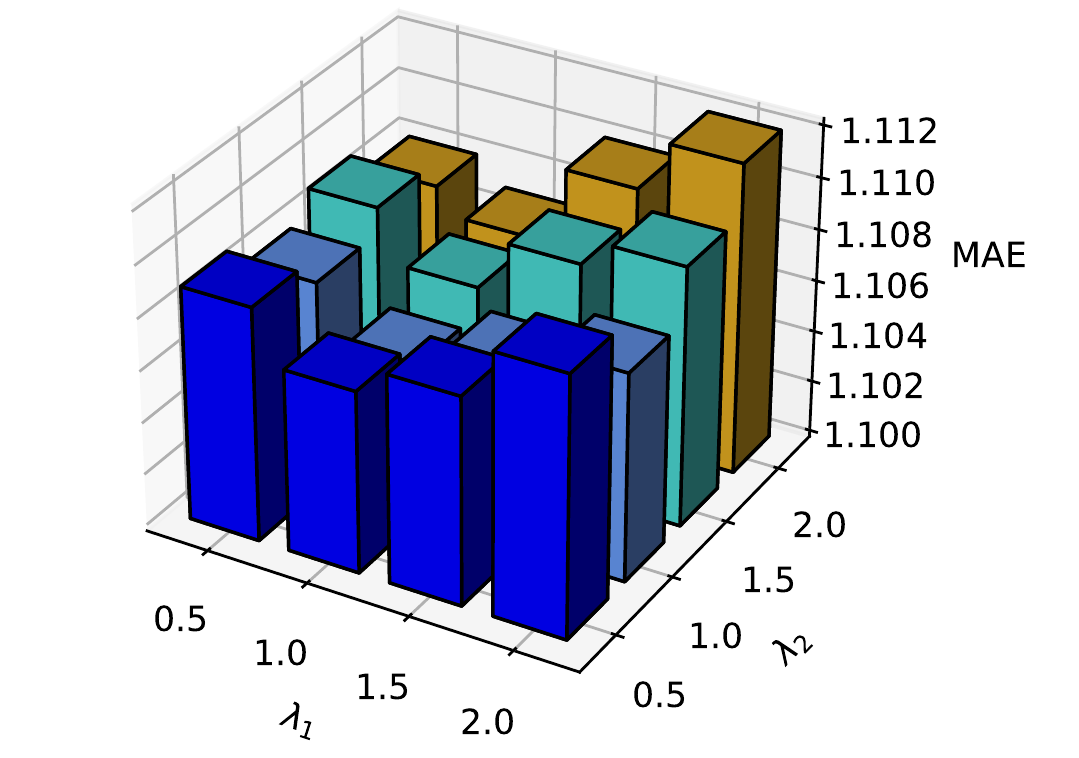}
	\caption{Ablation study of loss weights on CMU-MOSI in the incomplete modality setting.}
	\label{fig_mosi_ablation_loss_weights}
\end{figure}

After ablating the contribution of two feature restoration methods, we investigate the sensitivity of their corresponding weights in the loss function, i.e., $\lambda_1$ for low-level feature reconstruction and $\lambda_2$ for high-level feature attraction in Equation \ref{eq_loss_overall}.
We try several values for each weight parameter in the range of 0.5 to 2.0 with a step of 0.5.
The results of different combinations of $\lambda_1$ and $\lambda_2$ are shown in Fig. \ref{fig_mosi_ablation_loss_weights}. 
We mainly have the following observations: 1) All weight combinations have comparable performance, which suggests that our proposed method is insensitive to the  values of $\lambda_1$ and $\lambda_2$. 2) Too large weights may hurt the performance. 3) When $\lambda_1=1.0$ and $\lambda_2=1.0$, the model achieves the best performance. Thus, we use them as the default loss weights on this dataset.

\begin{figure}[t]
	\centering
	\includegraphics[width=\linewidth]{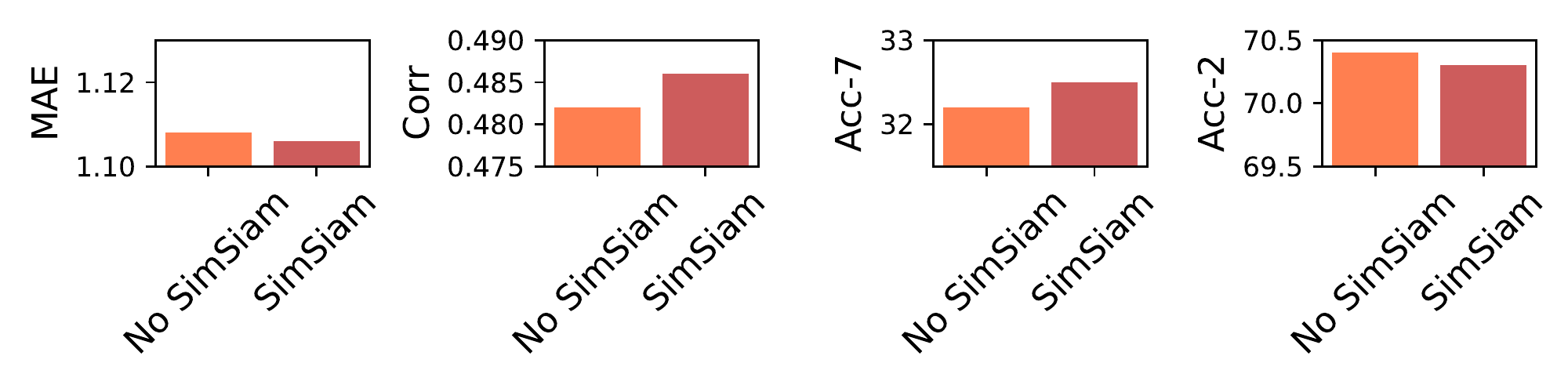}
	\caption{Ablation study of SimSiam for high-level feature attraction on CMU-MOSI in the incomplete modality setting.}
	\label{fig_mosi_ablation_simsiam}
\end{figure}

\begin{figure*}[t]
	\centering
	\includegraphics[width=\linewidth]{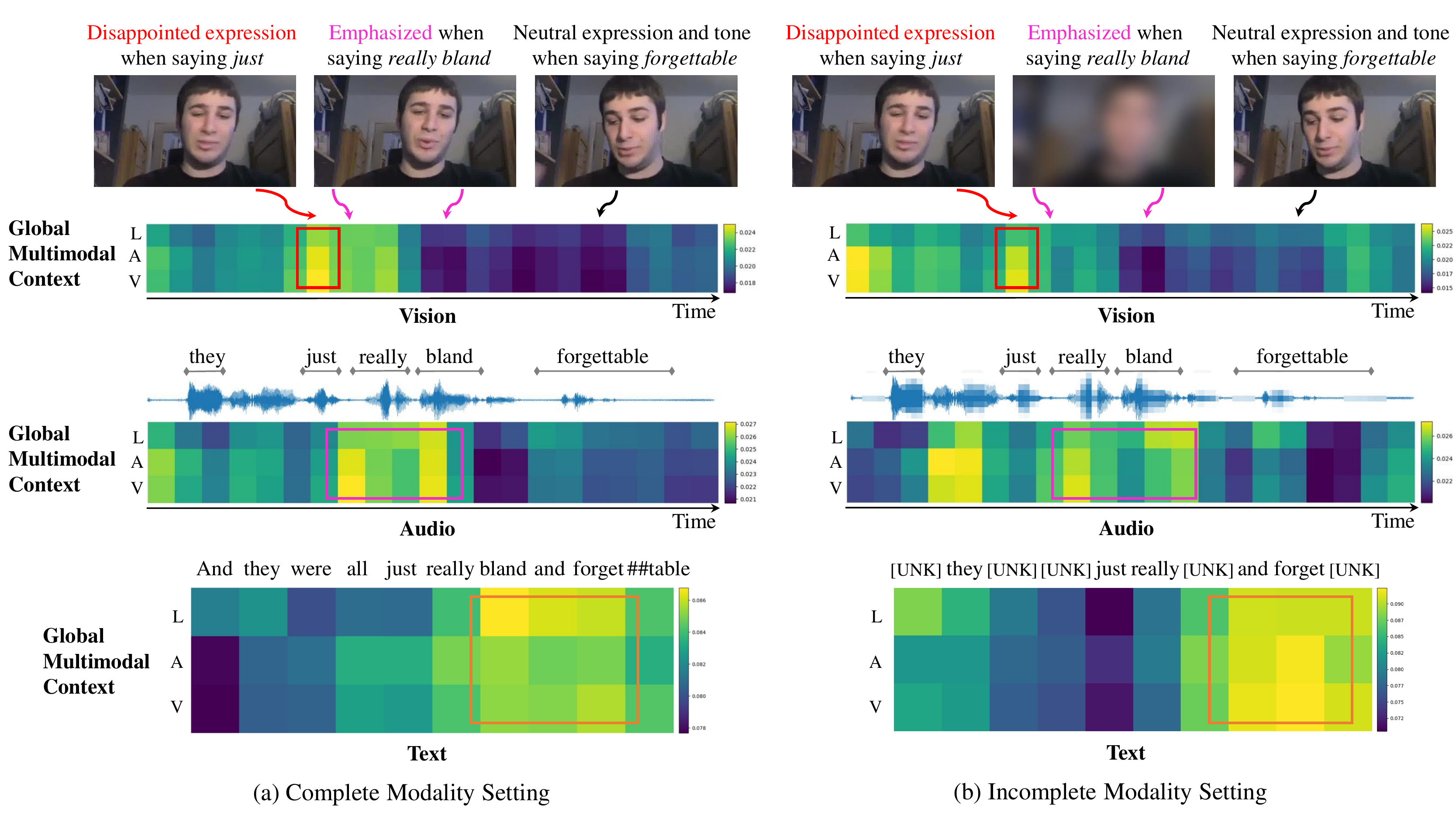}
	\caption{Visualization of global-local cross-modal attention weights from EMT in the complete and incomplete (under a missing rate of 0.5) modality setting. The sample jUzDDGyPkXU\_27 in the test set of CMU-MOSI is selected. For convenience, we only show the attention of the global multimodal context to local unimodal features (i.e., the global multimodal context is the target while each local unimodal feature is the source). High and meaningful attention areas are highlighted by colored rectangles. For both settings, we find that the global multimodal context has learned to pay attention to meaningful signals in each modality (e.g., disappointed facial expression, emphasized tone, and sentiment words).}
	\label{fig_vis_attn}
\end{figure*}

\begin{figure}[t]
	\centering
	\includegraphics[width=\linewidth]{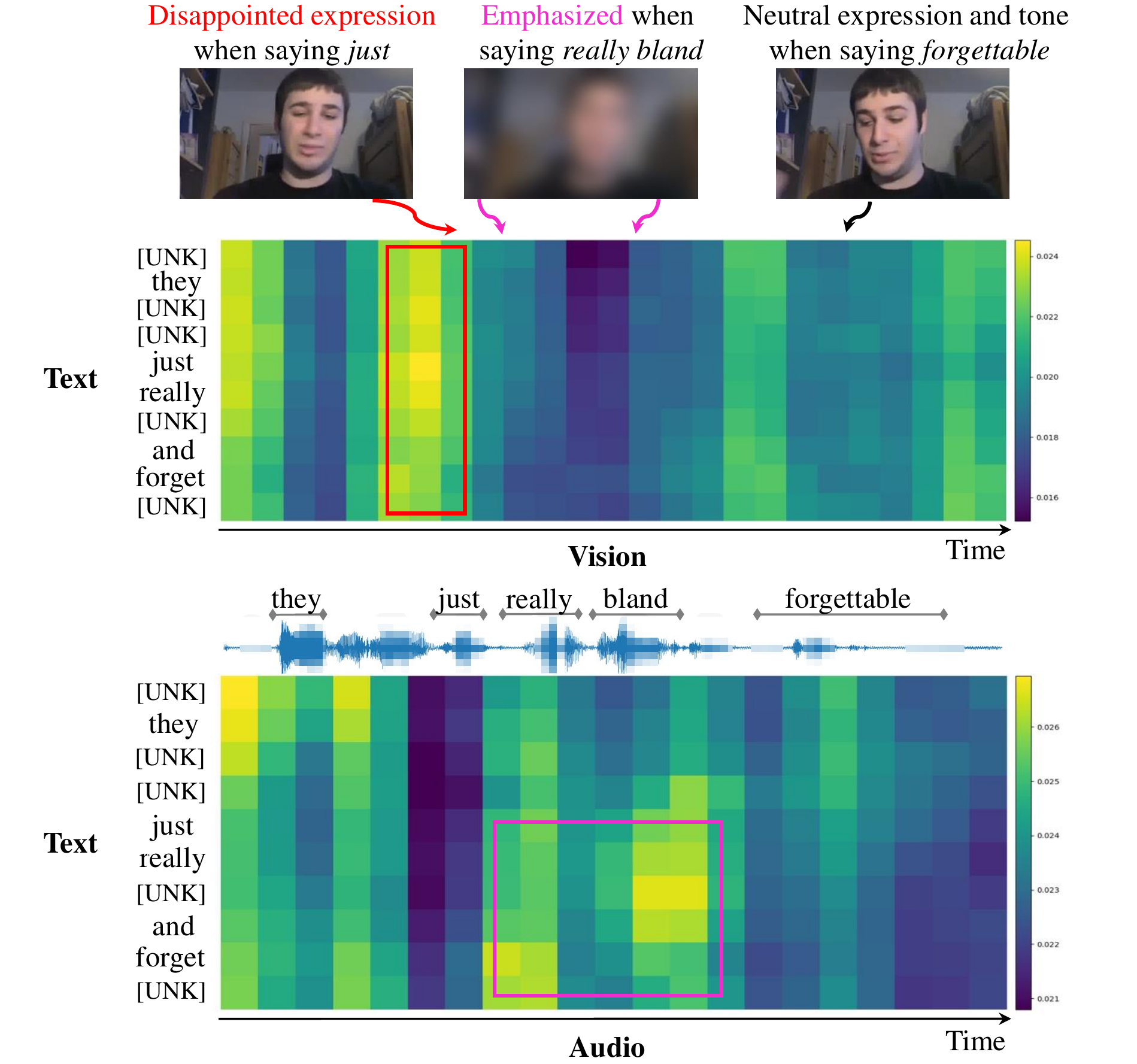}
	\caption{Visualization of local-local cross-modal attention weights from EMT (OOLL) in the incomplete modality setting (under a missing rate of 0.5). For convenience, we only show the attention of the text modality to audio and vision modalities. High and meaningful attention areas are highlighted by colored rectangles.}
	\label{fig_vis_attn_ooll}
\end{figure}

\subsubsection{SimSiam for High-Level Feature Attraction}
Finally, we analyze the impact of SimSiam in siamese representation learning for explicit high-level feature attraction. The alternative to SimSiam is to simply minimize the negative cosine similarity of utterance-level intra- and inter-modal representations between complete and incomplete views. We present the results of two methods in Fig. \ref{fig_mosi_ablation_simsiam}. We observe that the performance generally decreases when SimSiam is not used in siamese representation learning, which verifies the role of SimSiam in preventing the model from learning collapsed representations by virtue of its special architecture designs.

\subsection{Visualization Analysis}
To have a deeper understanding of how our model works when modeling unaligned multimodal sequences in both complete and incomplete modality settings, we empirically investigate the signals EMT captures by visualizing the cross-modal attention weights.
Fig. \ref{fig_vis_attn} presents the global-local cross-modal attention matrices of a sample selected from the test set of CMU-MOSI. For convenience, we only show the attention of the global multimodal context to each unimodal feature (i.e., $m \rightarrow g$, $m \in \{l, a, v\}$ in Equation \ref{eq_oagl}). Since the acoustic and visual sequences are too long (42 and 50 respectively), we use average pooling with a stride size of 2 to halve them. We also show the key video frames, audio waveform, text tokens from BERT, and alignment information for better interpretation.
It should be noted that, due to the segmentation error, the first word \textit{And} was not said by the speaker though it appears in the transcription provided by the dataset.

As shown in Fig. \ref{fig_vis_attn} (a), we find that our model captures meaningful global-local cross-modal interactions in the complete modality setting. Specifically, the global multimodal context pays its most attention to the visual segments where a disappointed facial expression occurs. For the audio modality, it learns to attend to the emphasized intervals in the acoustic sequence. While for the high-level text modality, we can see that sentiment words (e.g., \textit{bland} and \textit{forgettable}) are received stronger attention from the global multimodal context. These interesting observations qualitatively demonstrate the effectiveness of EMT for cross-modal interaction modeling. Finally, our model predicts a negative sentiment score of -2.1 for this sample, which is very close to the ground-truth label -2.0. 

In the incomplete modality setting, we use a missing rate of 0.5 to generate incomplete multimodal sequences. The missing audio and visual features are padded with zeros. The missing text tokens are replaced by the unknown token [UNK] in BERT. In Fig. \ref{fig_vis_attn} (b), we use blur background and mosaic to indicate the missing vision and audio modality features. 
It should be noted that they might not coincide with the real-generated temporal masks since they are only used for illustrative purposes.
From Fig. \ref{fig_vis_attn} (b), we observe that, although the attention matrices are more scattered due to the influence of random modality feature missing, our model still attends to those useful cross-modal signals captured in the complete modality setting. Notably, for the text modality, we find that the model pays more attention to the available token \textit{forget} as expected but also gives partial attention to the missing tokens (e.g., \textit{bland} and \textit{\#\#table}). Under such a moderate missing rate, our model prediction remains a negative score of -1.0. These results once again verify the efficacy of the proposed dual-level feature restoration to the improvement of model robustness.

In addition, we also visualize the local-local cross-modal attention weights to empirically investigate the differences between the proposed global-local fusion strategy (i.e., OAGL) and the previous local-local one (e.g., OOLL and OALL).
For simplicity, we only show OOLL in Fig. \ref{fig_vis_attn_ooll} as we have similar observations for OALL. 
We find that, although OOLL can capture a part of meaningful cross-modal correlation signals as OAGL, its attention matrices are typically low-rank (especially for the upper part in Fig. \ref{fig_vis_attn_ooll}), which demonstrates that a large amount of redundancy exists in local-local cross-modal interactions and it can be greatly reduced via the global-local interactions in OAGL to achieve efficient multimodal fusion. This empirical finding is also consistent with that in the previous study \cite{tsai2019multimodal}. Moreover, we notice that OOLL attends to more irrelevant cross-modal information than OAGL (such as the relatively higher attention to audio and vision modalities when saying \textit{forgettable}), implying that the local-local fusion strategy could also increase the risk of overfitting the spurious correlations in unaligned multimodal data.

\section{Conclusion}
In this paper, we have presented a generic and unified framework, named Efficient Multimodal Transformer with Dual-Level Feature Restoration (EMT-DLFR), for efficient and robust multimodal sentiment analysis. 
At the heart of EMT is the introduction of the global multimodal context, which enables effective and efficient exploration of global-local cross-modal interactions. It not only avoids the quadratic scaling cost of previous local-local cross-modal interaction modeling methods but also leads to performance gains. Furthermore, we utilize hierarchical parameter sharing to improve parameter efficiency and ease model training. To cope with random modality feature missing which typically occurs in realistic settings, DLFR employs both implicit low-level feature reconstruction and explicit high-level feature attraction to achieve robust representation learning from incomplete multimodal data. 
We find that, although the former is more effective than the latter, these two strategies are complementary to each other and thus can be combined to achieve better performance. 
Finally, extensive experiments on three datasets show that the proposed method achieves state-of-the-art performance in both complete and incomplete modality settings. In addition, empirical visualization analysis also demonstrates that our model can capture interpretable and robust cross-modal correlation signals for reliable sentiment prediction.

In future work, we hope to apply EMT-DLFR to the entire modality missing setting, noisy modality setting, and even more challenging noise-missing-mixed setting. It's also interesting to explore the effectiveness of the proposed method on different multimodal learning tasks and datasets. Besides, since we do not take into consideration the importance of each modality, it would be helpful to explicitly incorporate it into EMT-DLFR to further improve performance. Finally, although our EMT enjoys a linear scaling cost over the involved modalities, it still suffers from quadratic complexity with respect to the input sequence length. Thus, how to further reduce the complexity from $O(MT^2)$ to $O(MT)$ remains an attractive research direction.


%



\ifCLASSOPTIONcompsoc
  \section*{Acknowledgments}
\else
  \section*{Acknowledgment}
\fi


This work is supported by the National Natural Science Foundation of China (NSFC) (No.61831022, No.62276259, No.62201572, No.U21B2010), Beijing Municipal Science \& Technology Commission, Administrative Commission of Zhongguancun Science Park No.Z211100004821013, Open Research Projects of Zhejiang Lab (No.2021KH0AB06), and CCF-Baidu Open Fund (No.OF2022025).

\ifCLASSOPTIONcaptionsoff
  \newpage
\fi



\bibliographystyle{IEEEtran}
\bibliography{EMT-DLFR}
%



%

\begin{IEEEbiography}[{\includegraphics[width=1.1in,height=1.25in,clip,keepaspectratio]{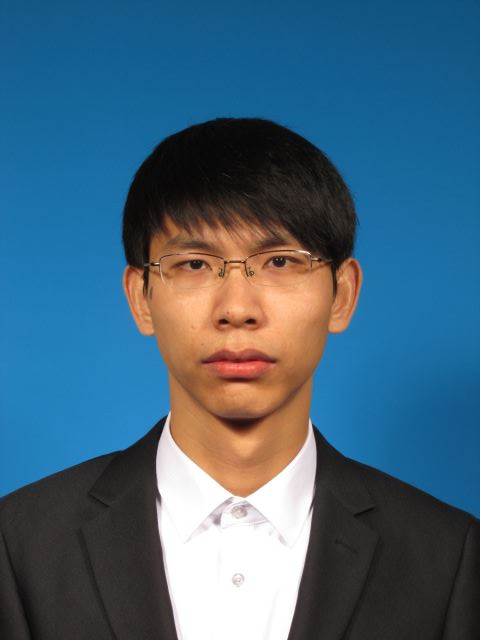}}]{Licai Sun}
received the B.S. degree from Beijing Forestry University, Beijing, China, in 2016, and the M.S. degree from University of Chinese Academy of Sciences, Beijing, China, in 2019. He is currently working toward the Ph.D. degree with the School of Artificial Intelligence, University of Chinese Academy of Sciences, Beijing,
China. His current research interests include affective computing, deep learning, and multimodal representation learning.
\end{IEEEbiography}

\begin{IEEEbiography}[{\includegraphics[width=1.1in,height=1.25in,clip,keepaspectratio]{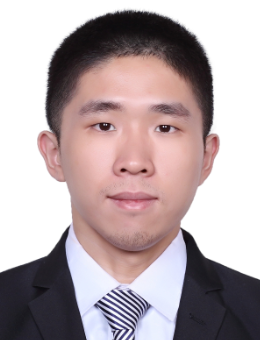}}]{Zheng Lian}
	received the B.S. degree from the Beijing University of Posts and Telecommunications (BUPT), Beijing, China, in 2016. And he received the Ph.D. degree from the Institute of Automation, Chinese Academy of Sciences, Beijing, China, in 2021. He is currently an Assistant Professor at National Laboratory of Pattern Recognition, Institute of Automation, Chinese Academy of Sciences, Beijing, China. His current research interests include affective computing, deep learning, and multimodal emotion recognition.
\end{IEEEbiography}

\begin{IEEEbiography}[{\includegraphics[width=1.1in,height=1.25in,clip,keepaspectratio]{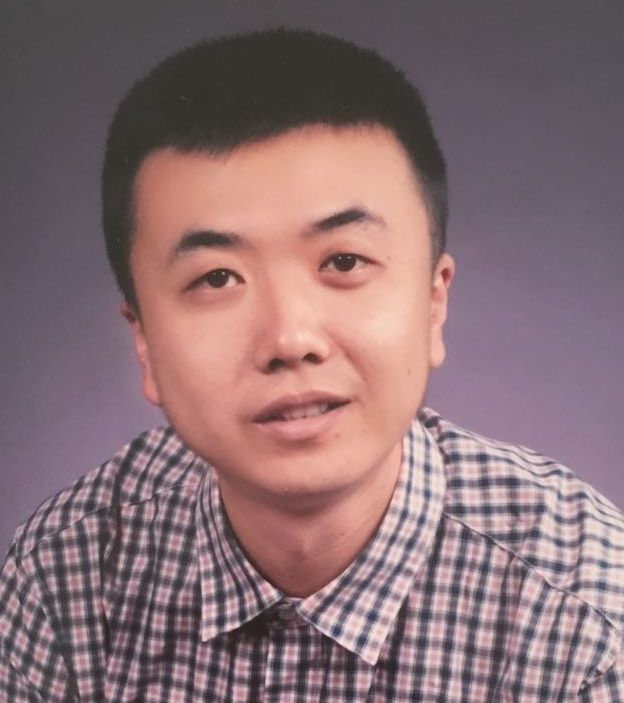}}]{Bin Liu}
	received his the B.S. degree and the M.S. degree from Beijing institute of technology (BIT), Beijing, China, in 2007 and 2009 respectively. He received Ph.D. degree from the National Laboratory of Pattern Recognition, Institute of Automation, Chinese Academy of Sciences, Beijing, China, in 2015. He is currently an Associate Professor in the National Laboratory of Pattern Recognition, Institute of Automation, Chinese Academy of Sciences, Beijing, China. His current research interests include affective computing and audio signal processing.
\end{IEEEbiography}

\begin{IEEEbiography}[{\includegraphics[width=1.1in,height=1.25in,clip,keepaspectratio]{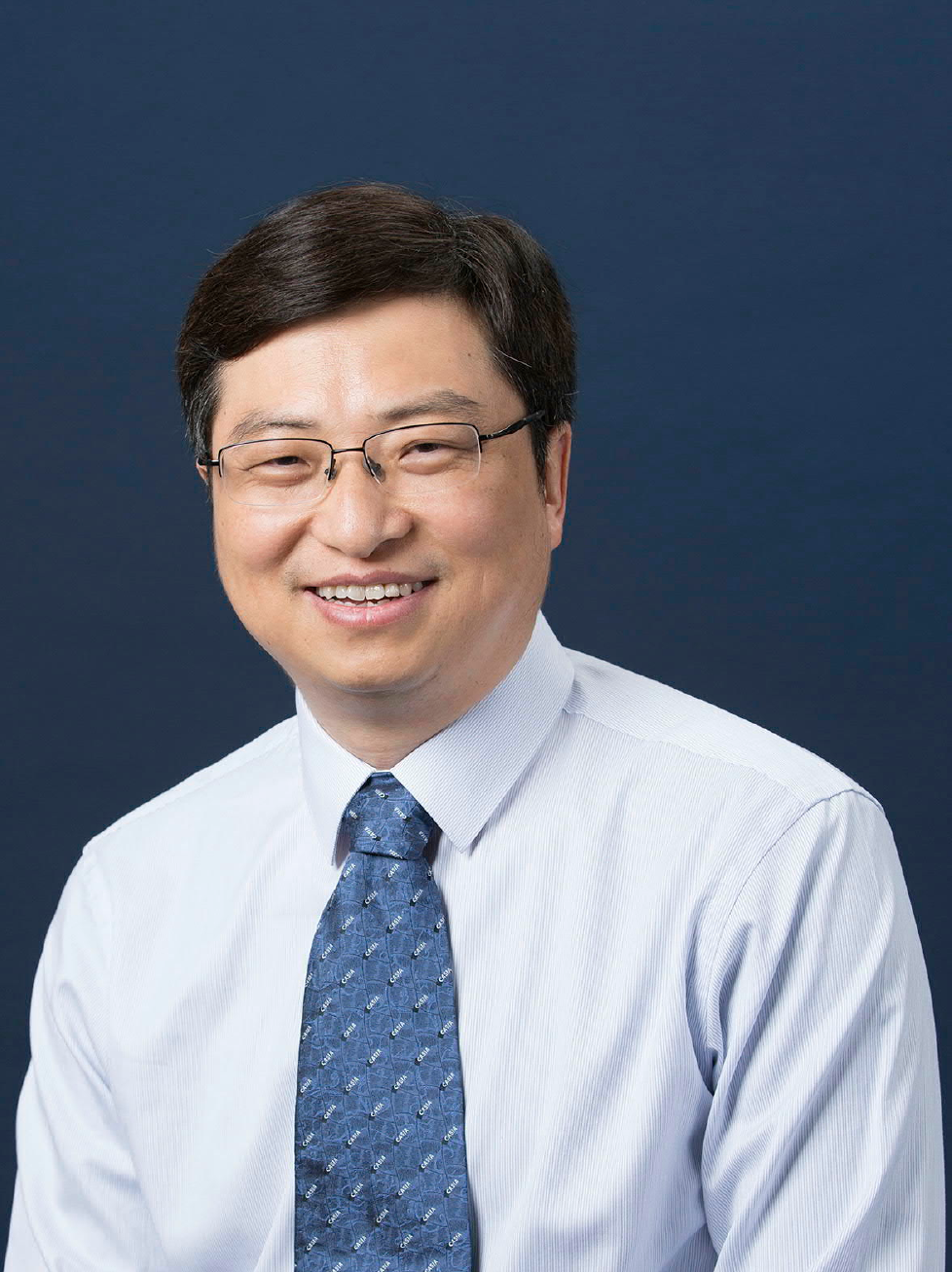}}]{Jianhua Tao}
    received the Ph.D. degree from Tsinghua University, Beijing, China, in 2001, and the M.S. degree from Nanjing University, Nanjing, China, in 1996. He is currently a Professor with Department of Automation, Tsinghua University, Beijing, China. He has authored or coauthored more than eighty papers on major journals and proceedings. His current research interests include speech recognition, speech synthesis and coding methods, human–computer interaction, multimedia information processing, and pattern recognition. He is the Chair or Program Committee Member for several major conferences, including ICPR, ACII, ICMI, ISCSLP, etc. He is also the Steering Committee Member for the IEEE Transactions on Affective Computing, an Associate Editor for Journal on Multimodal User Interface and International Journal on Synthetic Emotions, and the Deputy Editor-in-Chief for Chinese Journal of Phonetics.
\end{IEEEbiography}




\end{document}